\documentclass{article}

\usepackage[preprint]{neurips_2026}

\usepackage[utf8]{inputenc} 
\usepackage[T1]{fontenc}    
\usepackage{hyperref}       
\usepackage{url}            
\usepackage{booktabs}       
\usepackage{amsfonts}       
\usepackage{nicefrac}       
\usepackage{microtype}      
\usepackage{xcolor}         
\usepackage{subfigure}
\usepackage{graphicx}
\usepackage{multirow}
\usepackage{array}
\usepackage{graphicx}
\usepackage{microtype}
\usepackage{booktabs}
\usepackage{amssymb}
\usepackage{amsfonts}
\usepackage{amsmath}
\usepackage{mathrsfs}
\usepackage{multirow}
\usepackage{amssymb}
\usepackage{bbding}
\usepackage{subfigure} 
\usepackage{CJKutf8}
\usepackage{colortbl}
\usepackage{color}
\usepackage{booktabs}
\usepackage{makecell}
\usepackage{calligra}
\usepackage{algorithm} 
\usepackage{subcaption}
\usepackage{wrapfig}

\title{Do Large Language Models Plan Answer Positions? Position Bias in Multiple-Choice Question Generation}

\author{%
  Xuemei Tang \\
  The Hong Kong Polytechnic University \\
  \texttt{xuemeitang00@gmail.com}
  \And
  Xufeng Duan \\
  The Chinese University of Hong Kong \\
  \texttt{xufengduan@cuhk.edu.hk}
  \And
  Zhenguang G. Cai\thanks{Corresponding author.} \\
  The Chinese University of Hong Kong \\
  \texttt{zhenguangcai@cuhk.edu.hk}
}

\begin{document}

\maketitle

\begin{abstract}
Large language models (LLMs) are increasingly used to generate multiple-choice questions (MCQs), where correct answers should ideally be uniformly distributed across options. However, we observe that LLMs exhibit systematic position biases during generation.
Through extensive experiments with 10 LLMs and 5 vision-language models (VLMs) on three MCQ generation tasks, we show that these biases are structured, with similar patterns emerging within model families.
To investigate the underlying mechanisms, we conduct probing experiments and find that hidden representations in the question stem encode predictive signals of the correct answer position, suggesting that answer position may be implicitly planned during generation.
Building on this insight, we apply activation steering to manipulate internal representations and influence answer position. Our results show that steering can partially control positional preferences and substantially shift answer position distributions. Our findings provide a practical framework for studying implicit positional planning in LLMs and highlight the importance of controllable generation for reliable MCQ construction and evaluation.

\end{abstract}

\section{Introduction}

\begin{wrapfigure}{r}{0.4\textwidth}
\vspace{-5mm}
  \centering
  \includegraphics[width=0.4\textwidth]{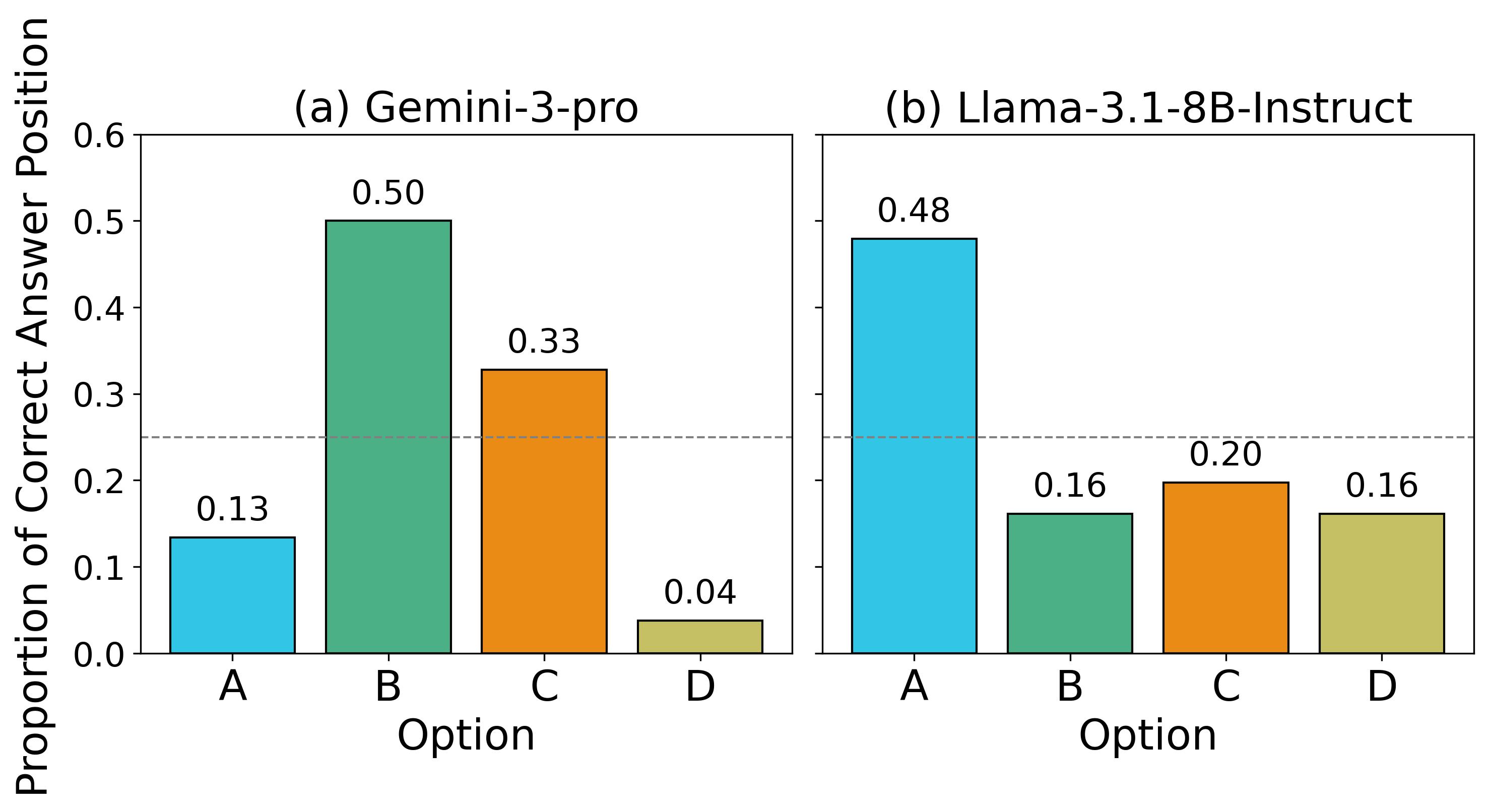}
  \caption{\textbf{Position bias in LLM-Generated Multiple Choice Questions.} 
  The gray dashed line at 25\% represents the theoretical unbiased baseline where correct answers are distributed evenly across all options.}
  \label{fig_intro}
  \vspace{-3mm}
\end{wrapfigure}

Large language models (LLMs) and visual-language models (VLMs) are increasingly used to generate multiple-choice questions (MCQs) for applications such as educational assessment~\citep{Mucciaccia_Meireles_Paixao_Wall_2025}, reading comprehension~\citep{Kalpakchi_Boye_2023}, and visual question-answering benchmark construction~\citep{Urailertprasert_Limkonchotiwat_Suwajanakorn_Nutanong_2024}. Ideally, the correct answers to generated questions should be uniformly distributed across answer options to ensure unbiased evaluation. However, in practice, LLMs exhibit systematic position biases, disproportionately favoring certain answer positions.

As illustrated in Figure~\ref{fig_intro}, this behavior varies substantially across models. For example, Gemini-3-pro exhibits a pronounced \textit{center bias}, preferring options ``B'' and ``C'', whereas Llama-3.1-8B-Instruct demonstrates a strong \textit{primacy bias}, heavily skewing toward option ``A''. These patterns suggest that position bias is systematic and tied to model-specific generation behavior.

While prior work has investigated selection bias in multiple-choice question answering, showing that LLMs tend to favor particular option identifiers (e.g., ``Option A’’)~\citep{Pezeshkpour_Hruschka_2024,Zheng_Zhou_Meng_Zhou_Huang_2024,Loginova_Bezrukov_Shekhar_Kravets_2025}, it remains unclear whether similar biases arise during the generation process itself. In particular, it is unknown whether LLMs exhibit consistent positional preferences when generating correct answers, and whether such preferences are driven by internal representations rather than surface-level biases.

Beyond this line of work, recent studies suggest that LLMs may engage in implicit planning during generation~\citep{Pochinkov_Benoit_Agarwal_Majid_Ter-Minassian_2024, Maar_Paperno_McDougall_Nanda_2026}. For example, hidden representations have been shown to encode information about future outputs beyond the next token~\citep{Dong_Zhou_Liu_Yang_Lu_2025}, and interventions on hidden states prior to generation can influence subsequent outputs, including intermediate tokens and final answers~\citep{Maar_Paperno_McDougall_Nanda_2026}.
From this perspective, MCQ generation can be viewed as a structured decision-making process, where determining the position of the correct answer is part of the generation strategy. This suggests that position biases may not be purely incidental, but instead reflect latent planning mechanisms encoded within the model.

Building on this intuition, we ask three key questions: (i) Do LLMs systematically exhibit position bias in MCQ generation across different model families? (ii) Does this bias reflect implicit planning mechanisms during generation? and (iii) Can such behavior be controlled or mitigated through targeted intervention at decoding time?

To answer these questions, we first conduct a systematic analysis of position bias across LLMs and VLMs on three multiple-choice question generation tasks, showing consistent preferences within model families. Next, we probe internal representations and find that answer position can be partially predicted from question token representations, suggesting that positional information may already be encoded during generation.
Finally, we demonstrate that activation steering (mean-difference and classifier-based methods) can influence answer position, establishing a causal link between internal representations and position bias.



\section{Related Work}



\paragraph{Selection and position bias in LLMs}

Prior work has shown that LLMs exhibit selection bias when answering MCQs, often preferring specific option identifiers such as ``Option A'' due to token-level probability biases~\citep{Zheng_Zhou_Meng_Zhou_Huang_2024, Guda_Francis_Ashungafac_Joe-Wong_Busogi_2025}. 
This bias frequently manifests as positional preferences, where models disproportionately select answers appearing in certain positions. For example, GPT-4 has been observed to favor answers in the first position~\citep{Zheng_Chiang_Sheng_Zhuang_Wu_Zhuang_Lin_Li_Li_Xing_et_2023}. 
Similar ordering effects have been reported in other settings, including surveys and bilingual tasks, where model responses are influenced by option order and labeling rather than underlying semantics~\citep{Dominguez-Olmedo_Hardt_Mendler-Dunner_2024,Li_Li_Xiang_Liu_Deng_Garcia_2024,Wei_Wu_Huang_Chen_2024}. While these studies establish that LLMs are sensitive to option ordering during answer selection, they predominantly focus on the inference stage. In contrast, we study position bias in the \emph{generation} process, asking whether LLMs exhibit systematic preferences when \emph{constructing} MCQs, specifically in determining the position of the correct answer.

Regarding evaluation metrics for selection or position bias, \citet{Choi_Xu_Xue_Eckman_Reddy_2025} introduced Choice Kullback–Leibler Divergence (CKLD), which evaluates the discrepancy between the predicted answer distribution and the ground-truth distribution. 
Complementary to distributional measures, other metrics assess variability across answer options. For instance, Relative Standard Deviation (RSD)~\citep{Reif_Schwartz_2024} and Standard Deviation of Recall (RStd)~\citep{Zheng_Zhou_Meng_Zhou_Huang_2024} capture the dispersion in per-option accuracy and recall, respectively.

\paragraph{Planning Mechanisms in LLMs}
Recent studies suggest that LLMs exhibit various forms of planning during generation.
\cite{Wu_Morris_Levine_2024} proposed pre-caching and breadcrumb mechanisms, where models compute features in advance that later support future predictions.
\cite{Men_Cao_Jin_Chen_Liu_Zhao_2024} and \cite{Pochinkov_Benoit_Agarwal_Majid_Ter-Minassian_2024} demonstrated that internal representations encode information about short-term and paragraph-level future content, indicating look-ahead behavior beyond token-level prediction.
More broadly, \cite{Dong_Zhou_Liu_Yang_Lu_2025, Maar_Paperno_McDougall_Nanda_2026} found that hidden states encode global properties of future outputs and exhibit planning-like behavior across diverse tasks.
Overall, these findings suggest that hidden representations carry information about future outputs, providing a potential basis for downstream manipulation such as activation steering.

\paragraph{Activation Steering}
Recent work has shown that LLM behavior can be controlled through activation-level interventions on hidden representations~\citep{Loula_LeBrun_Du_Lipkin_Pasti_Grand_Liu_Emara_Freedman_Eisner_et_al_2025, Sun_Peng_Dai_Bai_Cao_2025}. Prior studies propose methods such as latent steering vectors, Activation Addition (ActAdd), and Contrastive Activation Addition (CAA), which steer generation by adding contrastive directions to hidden states without fine-tuning~\citep{Subramani_Suresh_Peters_2022,Turner_Thiergart_Leech_Udell_Vazquez_Mini_MacDiarmid_2023,Rimsky_Gabrieli_Schulz_Tong_Hubinger_Turner_2024}. Beyond behavioral control, another line of work studies the gap between internal representations and model outputs. Inference-Time Intervention (ITI) has been proposed to mitigate this mismatch by steering activations toward more faithful outputs without retraining~\citep{Li_Patel_Viegas_Pfister_Wattenberg_2024}.

More recent work extends activation steering to more complex and compositional settings. \cite{Stolfo_Balachandran_Yousefi_Horvitz_Nushi_2025} presented that instruction-derived activation vectors enable modular and transferable control over instruction following, while Steer2Adapt introduces reusable semantic subspaces with optimized steering coefficients for stable and data-efficient control across tasks~\citep{Han_Xu_Xuan_Song_Ouyang_Tian_Jiang_Qian_Jiang_Sun——al_2026}.

Overall, these studies suggest that LLM behavior can be effectively controlled through targeted manipulation of internal representations without modifying model parameters.


\section{Method}
Our framework consists of three components: (1) a controlled MCQ generation setup for measuring position bias, (2) probing analyses of hidden representations for answer position prediction, and (3) activation intervention experiments for causal manipulation of answer positions.
\subsection{MCQ Generation Task Design}
~\label{mcq_des}
To study position bias in MCQ generation, we design a controlled setup where LLMs generate questions with four answer options and an explicitly specified correct answer. We standardize the answer format (e.g., options A–D) and ensure that question semantics do not constrain the correct answer position, allowing us to attribute observed biases to the model rather than task artifacts. 

We consider three classical multiple-choice generation scenarios as our test bed: Knowledge-based MCQ (K-MCQ), Reading Comprehension MCQ (RC-MCQ), and Vision-Language MCQ (VL-MCQ).


\textbf{Task 1: K-MCQ generation.} The first experimental setting focuses on educational assessment. 
We select five educational disciplines (Mathematics, Chemistry, History, Geography, and Biology) as the background domains. 
For each discipline, we collect five key examination knowledge points from an educational website\footnote{https://www.zxxk.com/}. 
We then prompt LLMs to generate 50 multiple-choice question items for each knowledge point, 
where each item consists of a question stem, four answer options, and one correct answer option ID. 
In total, we construct 200 question–answer pairs for each discipline. 

\textbf{Task 2: RC-MCQ generation.}
The second experimental setting focuses on reading comprehension. We collect 30 news articles from the BBC News website (dated December 5, 2025), which serve as reading passages. Based on these passages, we prompt LLMs to generate 16 question–answer pairs per article, yielding a total of 480 question–answer pairs.

\textbf{Task 3: VL-MCQ generation.}
The third experimental setting focuses on visual question answering (VQA). We randomly sample 30 images from the COCO\_test\_2015 dataset. To mitigate potential data leakage, we first use gemini-3-pro-preview to generate textual descriptions of the images, and then use gemini-3-pro-image-preview to generate abstract images based on these descriptions. Finally, we prompt VLMs to generate 16 multiple-choice question–answer pairs for each abstract image, resulting in a total of 480 question–answer pairs.

We determine the number of generated samples via a pilot study to ensure stable and statistically reliable estimation of position bias.

\subsection{Probing for Answer Position Planning}
\label{method_2}

To investigate whether position bias arises from internal planning, we probe hidden representations to predict the correct answer position.

Formally, let $x = (x_1, \ldots, x_m)$ denote the tokenized question stem. For an $L$-layer decoder-only LLM, let $H_x^{(l)} = (H_{x}^{(l,1)}, \ldots, H_{x}^{(l,m)})$ denote the hidden representations at layer $l \in [0, L]$. The target attribute is the answer position index $y \in \{1, \ldots, K\}$.

We train a lightweight classifier $f_\theta$ on token-level hidden states, where each $H_{x}^{(l,i)}$ is used to predict $y$:
\begin{equation}
\hat{y} = f_\theta(H_{x}^{(l,i)})
\end{equation}

We evaluate probing performance across layers and token positions to identify where predictive signals are most salient. Performance significantly above a random baseline indicates that information about answer position is encoded in stem representations prior to option generation, suggesting early planning of answer position.

\subsection{Intervention}
\label{method_3}

If a model performs implicit forward planning, interventions applied at earlier positions may influence outputs generated several tokens later~\citep{Maar_Paperno_McDougall_Nanda_2026}. We investigate whether answer positions can be manipulated via activation-level steering on question stems.

We construct steering vectors using two methods:

\textbf{Mean-Difference.}
We compute a direction in representation space by taking the difference between mean hidden states associated with target and source answer positions:

\begin{equation}
v_{src \rightarrow tgt}^{(l,i)} 
= \frac{1}{|\mathcal{H}_{tgt}^{(l,i)}|}
\sum_{H \in \mathcal{H}_{tgt}^{(l,i)}} H
\;-\;
\frac{1}{|\mathcal{H}_{src}^{(l,i)}|}
\sum_{H \in \mathcal{H}_{src}^{(l,i)}} H
\end{equation}

where, $\mathcal{H}_{src}^{(l,i)} = \{ H_{x}^{(l,i)} \mid \text{Answer}(H_{x}^{(l,i)}) = src \}$ and $\mathcal{H}_{tgt}^{(l,i)} = \{ H_{x}^{(l,i)} \mid \text{Answer}(H_{x}^{(l,i)}) = tgt \}$ represent the hidden states that associated answer $src$ and $tgt$. 
$v_{src \rightarrow tgt}^{(l,i)}$ represents the average shift in representation space from favoring option $src$ to favoring option $tgt$.

\textbf{Classifier-Based.}
We train an $\ell_2$-regularized logistic regression classifier to distinguish source and target positions, and use its weight vector as the steering direction:
\begin{equation}
v_{src \rightarrow tgt}^{(l,i)} = w^{(l,i)}.
\end{equation}
Here, $w^{(l,i)}$ defines a discriminative direction orthogonal to the decision boundary in representation space.

Full implementation details are provided in Appendix~\ref{appendix:steering_vectors}.

\paragraph{Steering Mechanism.}
We apply intervention at token position $i$ in layer $l$ by modifying hidden states as:
\begin{equation}
\tilde{H}_x^{(l,i)} = H_x^{(l,i)} + \alpha \cdot v_{src \rightarrow tgt}^{(l,i)},
\end{equation}
where $\alpha$ controls intervention strength. The model then continues generation to produce answer options and the final correct answer identifier.

\section{Position Bias Experiments}

\subsection{Experimental Setup}
\label{model_selection}
We evaluate position bias across a diverse set of 10 LLMs on Task 1 and Task 2, including both open-source and closed-source systems. 
Open-source LLMs include\textit{ Mistral-7B-Instruct-v0.3, Llama-3.1-8B-Instruct, Llama-3.2-3B-Instruct, Qwen3-4B-Instruct-2507, Qwen3-30B-Instruct-2507, and Qwen3-30B-Thinking-2507}. 
Closed-source LLMs include \textit{DeepSeek-Reasoner, DeepSeek-Chat, Gemini-2.5-Flash, and Gemini-3-Pro}.
For Task 3, we evaluate five VLMs, including four open-source models, \textit{Llama-3.2-11B-Vision-Instruct, LLaVA-v1.6-Mistral-7B, Qwen3-VL-235B-Instruct, and Qwen3-VL-235B-Thinking}, and one closed-source model, \textit{Gemini-3-Pro}. The model implementation details are described in Appendix~\ref{exp1_setup}.

For each model, we report the distribution of correct answer positions. We further evaluate two bias metrics (see Appendix~\ref{bias_m}) across the three tasks, and conduct a Chi-square test for statistical significance (see Appendix~\ref{metrics_results}).

\begin{figure}[h]
 \vspace{-3mm}
    \centering
   \subfigure[Task 1: K-MCQ\label{}]{\includegraphics[width=0.35\linewidth]{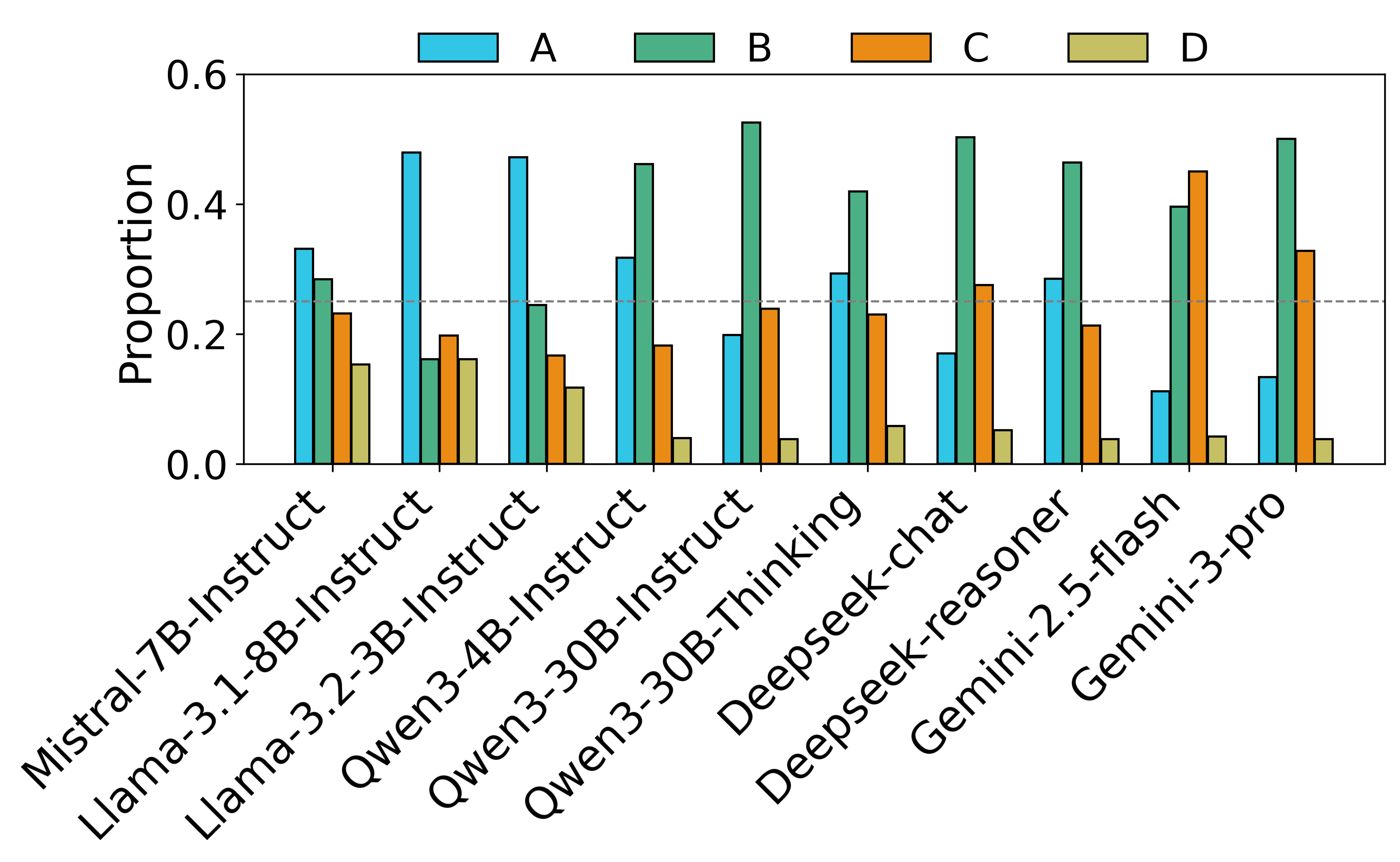}}
    \subfigure[Task 2: RC-MCQ\label{}]{\includegraphics[width=0.35\linewidth]{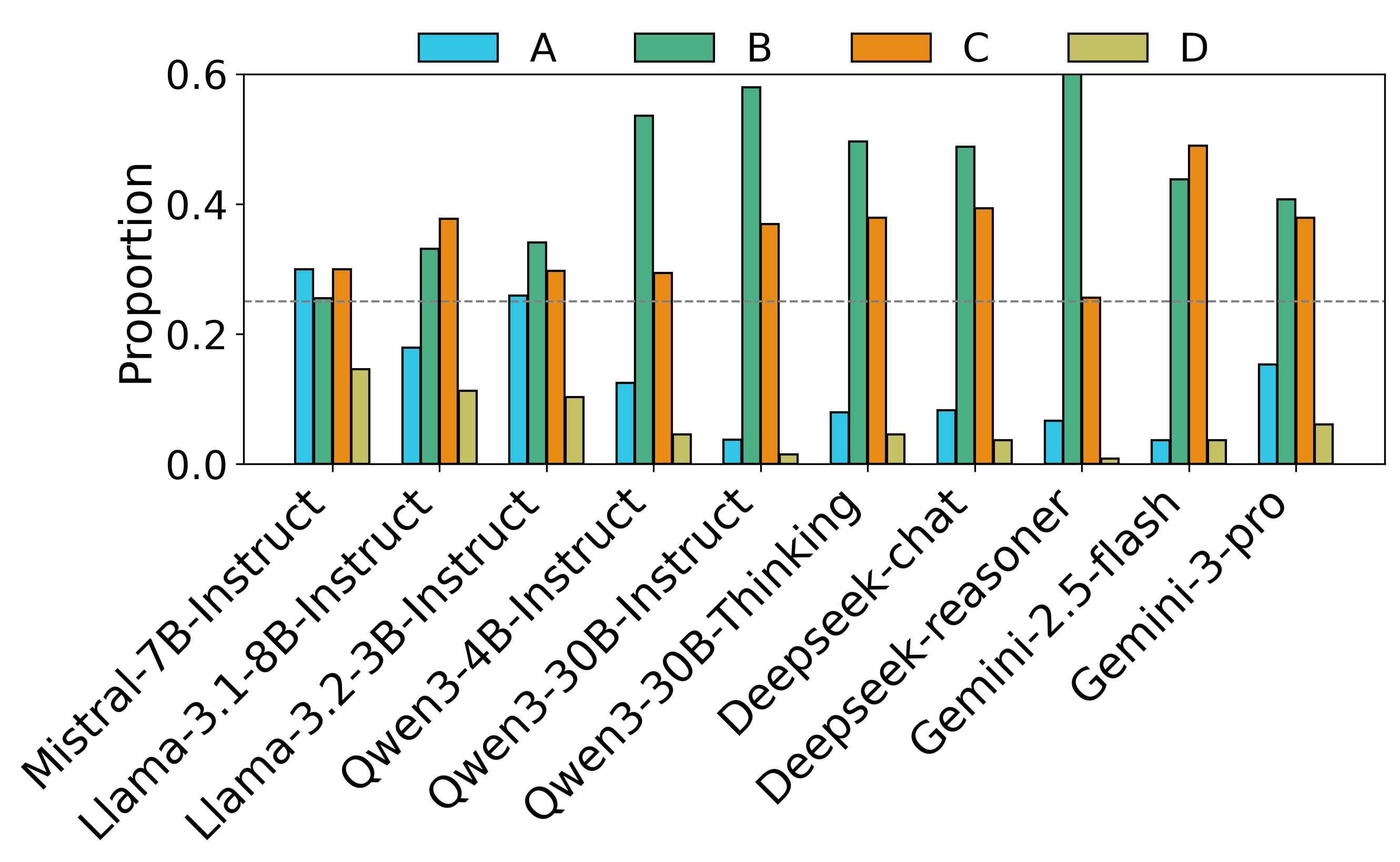}}
    \subfigure[Task 3: VL-MCQ\label{}]{\includegraphics[width=0.24\linewidth]{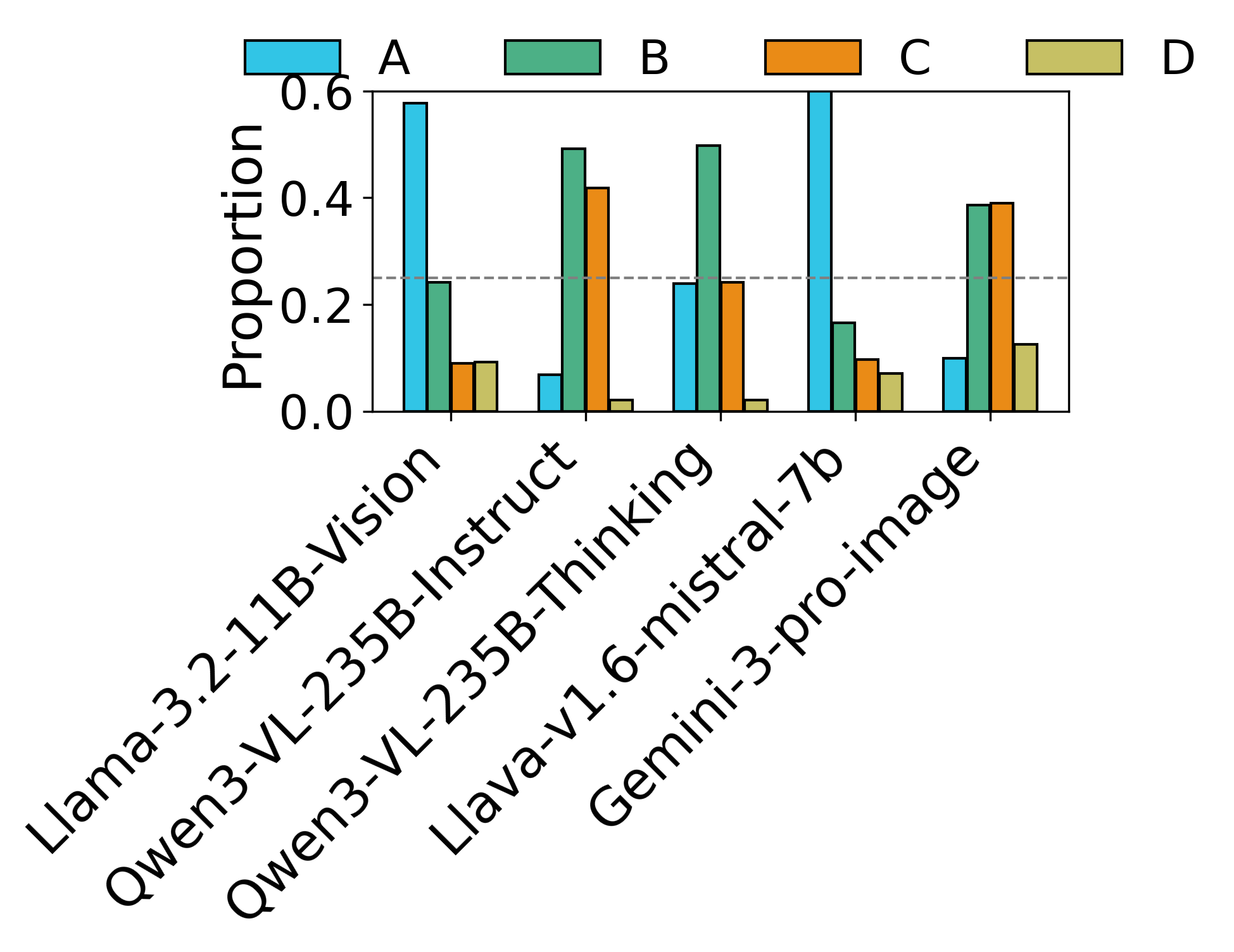}}
    \caption{Distribution of correct answer positions across three MCQ generation tasks. }
    \label{task1-3_distribution}
    \vspace{-3mm}
\end{figure}

\subsection{Position Bias Results}
First, we visualize the distribution of correct answer positions across all three tasks and models, as shown in Figure~\ref{task1-3_distribution}. We observe consistent and structured positional biases across model families.

LLaMA models consistently favor option ``A'' in K-MCQ and VL-MCQ settings. In contrast, Qwen3 and DeepSeek variants tend to prefer option ``B'', while Gemini models more frequently assign correct answers to options ``B'' and ``C''. Across nearly all models and tasks, option ``D'' is systematically underrepresented.


\subsection{Effect of Option Identifiers on Position Bias}

\begin{wrapfigure}{l}{0.6\textwidth}
\vspace{-3mm}
\centering
    \includegraphics[width=0.6\textwidth]{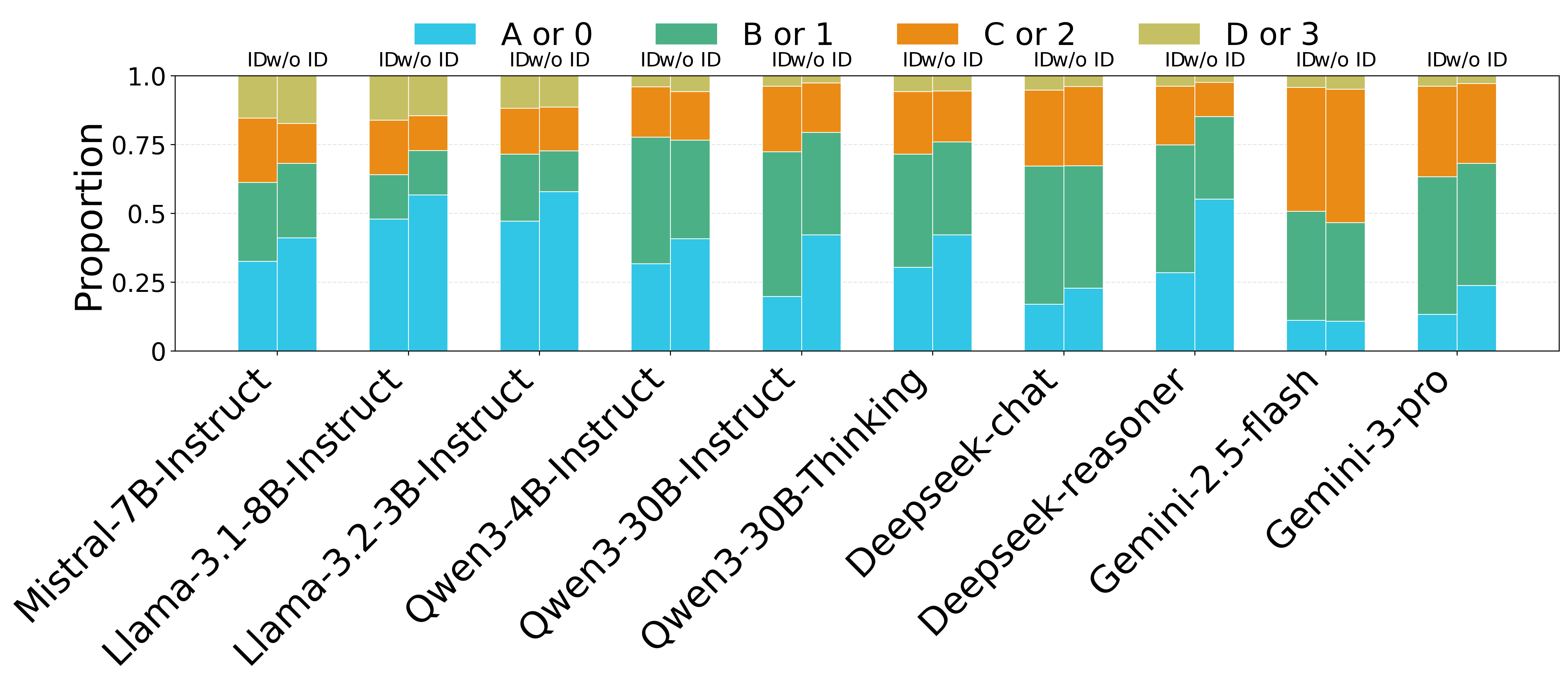}
\caption{
Comparison of correct answer position distribution with and without option identifiers on Task 1. 
}
\label{Task1_normal_ID_woID}
\vspace{-3mm}
\end{wrapfigure}

We investigate whether position bias is driven by explicit option identifiers (e.g., A/B/C/D). To this end, we remove option labels and prompt models to generate four candidate answers along with the correct answer text without explicit ordering markers.

This identifier-free setting eliminates surface-level positional cues, allowing us to isolate implicit ordering preferences.

Figure~\ref{Task1_normal_ID_woID} compares answer position distributions with and without identifiers for Task 1 (with full results for Tasks 2 and 3 in Appendix~\ref{Appendix_ID}). We observe a consistent shift toward the first position across models in the identifier-free setting.

These findings suggest that option identifiers provide structured positional cues that help regulate answer-position selection. Statistical analyses in Appendix~\ref{Appendix_ID_stats} further show that the increased first-position preference remains significant after controlling for model-specific variability, indicating that explicit identifiers systematically mitigate primacy bias. However, removing identifiers does not eliminate positional bias entirely, as many models still exhibit non-uniform answer distributions, with a substantially stronger first-position preference.



\begin{wrapfigure}{l}{0.6\textwidth}
\vspace{-4mm}
\centering
    \includegraphics[width=0.6\textwidth]{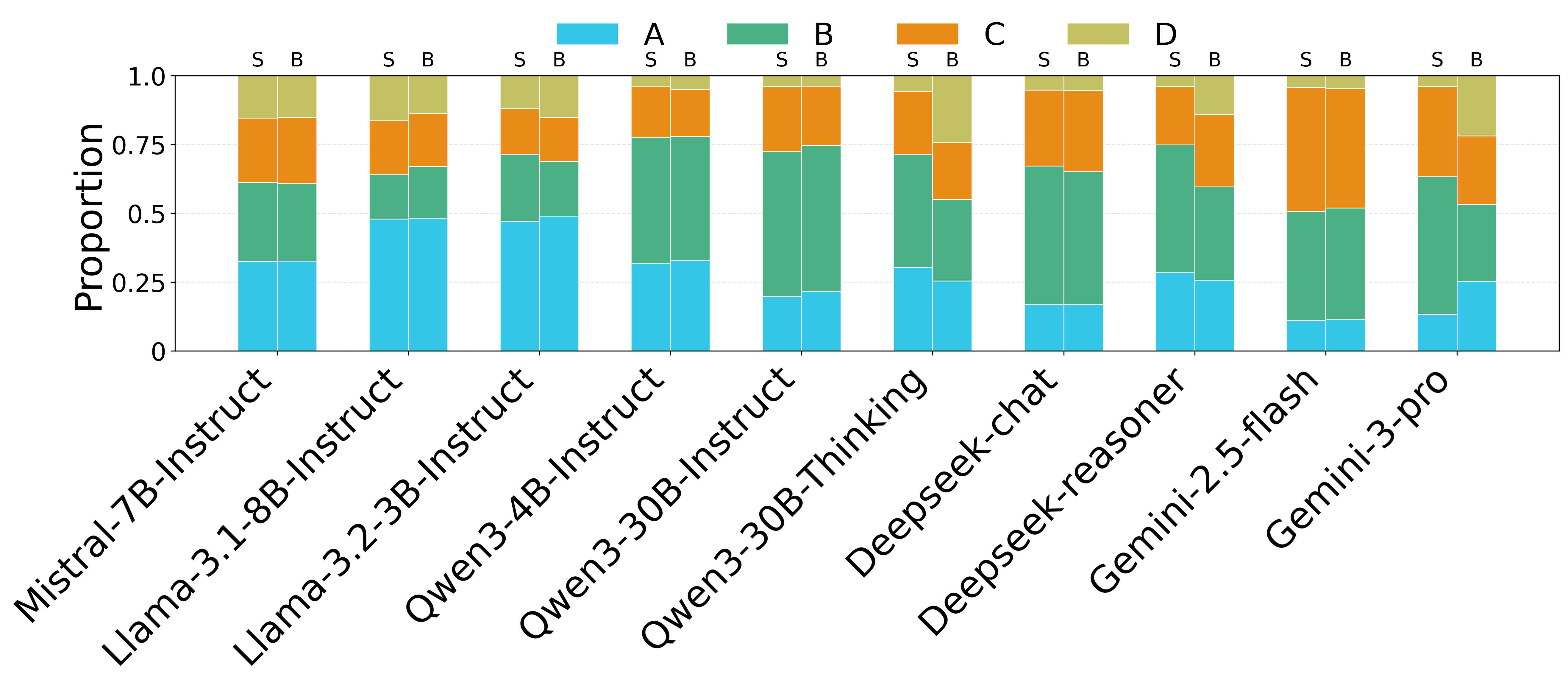}
\caption{Comparison of standard and balanced prompting strategies with respect to position bias in Task 1. ``S'' denotes the standard prompt, and ``B'' denotes the balanced prompt. }
\label{Task1_normal_balance}
\vspace{-4mm}
\end{wrapfigure}

\section{Can Rule-based Prompting Mitigate Position Bias?}

We investigate whether position bias can be mitigated via simple rule-based prompting. Specifically, we add an explicit instruction:

\textit{“Ensure the distribution of correct answer options (A/B/C/D) is approximately balanced across the {count} questions.”}

We refer to this as \textbf{balanced prompting}, which explicitly encourages uniform answer distributions. We compare it against standard prompting.

As shown in Figure~\ref{Task1_normal_balance} (Task 1; Tasks 2–3 in Appendix~\ref{Appendix_balance}), balanced prompting consistently yields more uniform answer distributions. \textbf{Reasoning-oriented models benefit the most}: DeepSeek-Reasoner and Gemini-3-Pro show near-zero bias on Tasks 2–3 (Appendix~\ref{Appendix_balance}), while Qwen3-30B-Thinking shows substantial reductions across all tasks. However, improvements vary by model, \textbf{larger or reasoning-enhanced models respond well, whereas smaller models still struggle.} Overall, prompt-level constraints can mitigate position bias, but their effectiveness depends on the model’s ability to follow global distributional instructions.

\section{Do LLMs Implicitly Plan Answer Positions?}

\subsection{Probing Position Information}

\begin{figure}
\vspace{-1mm}
\centering
\subfigure[Average F1]{
    \includegraphics[width=0.35\textwidth]{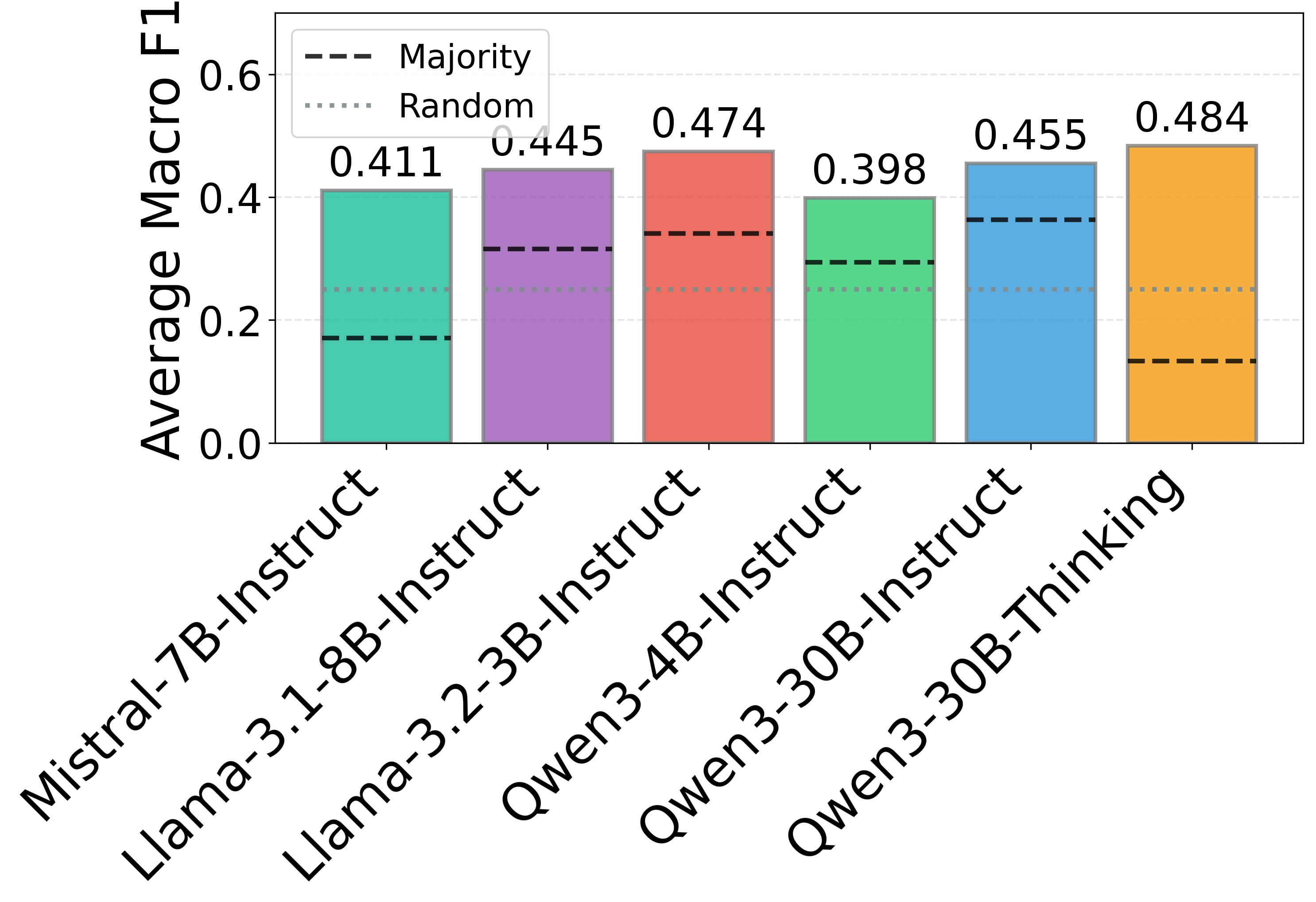}
}
\subfigure[Peak F1]{
    \includegraphics[width=0.35\textwidth]{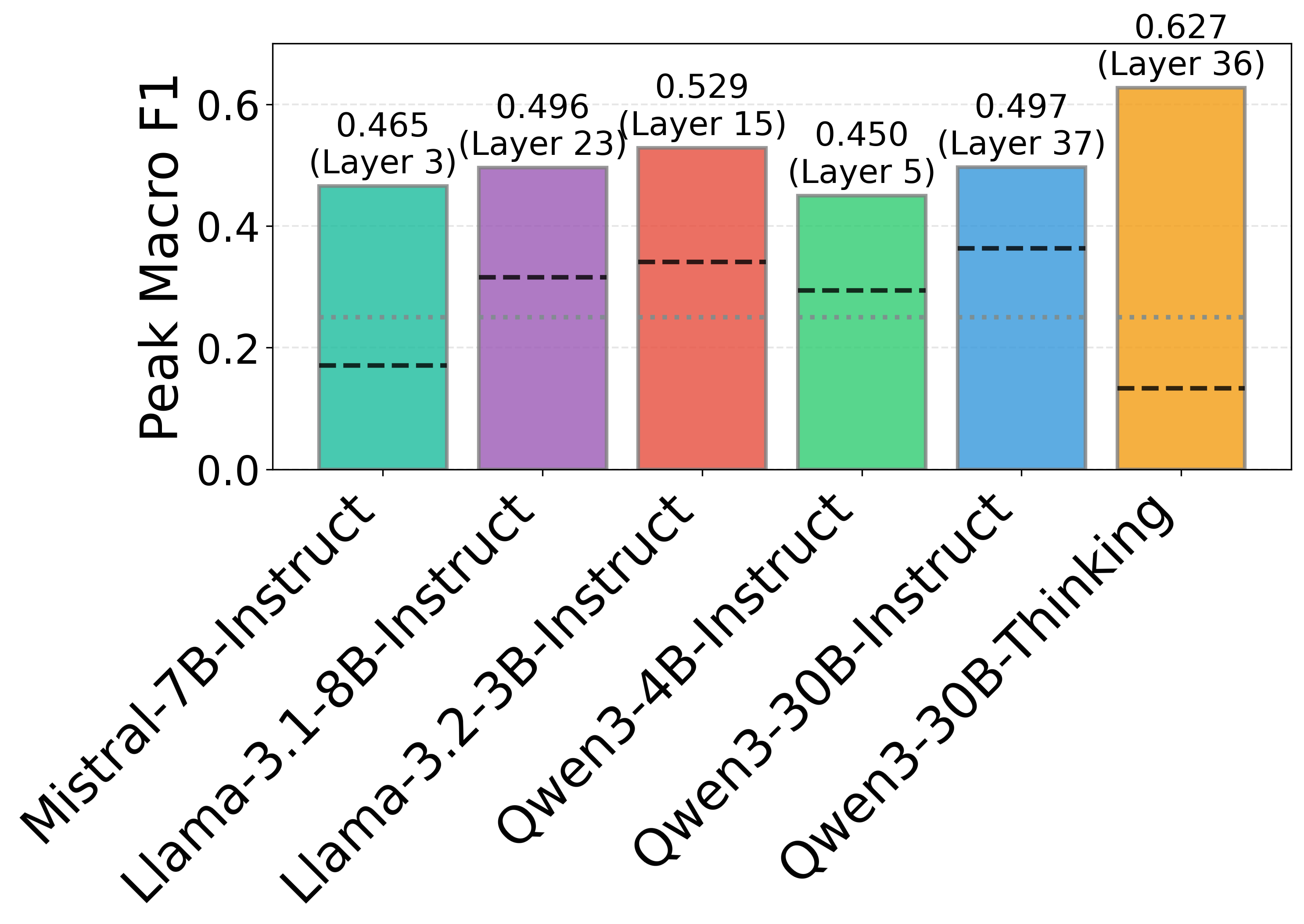}
}
\caption{Average macro F1 of all layers and peak macro F1 scores of MLP probes for predicting correct answer positions. ``Majority'' refers to the majority-class baseline, while ``Random'' corresponds to uniform random guessing over options A–D.}
\vspace{-3mm}
\label{aver_peak}
\end{figure}

\begin{figure}
\centering
\vspace{-3mm}
\includegraphics[width=0.9\textwidth]{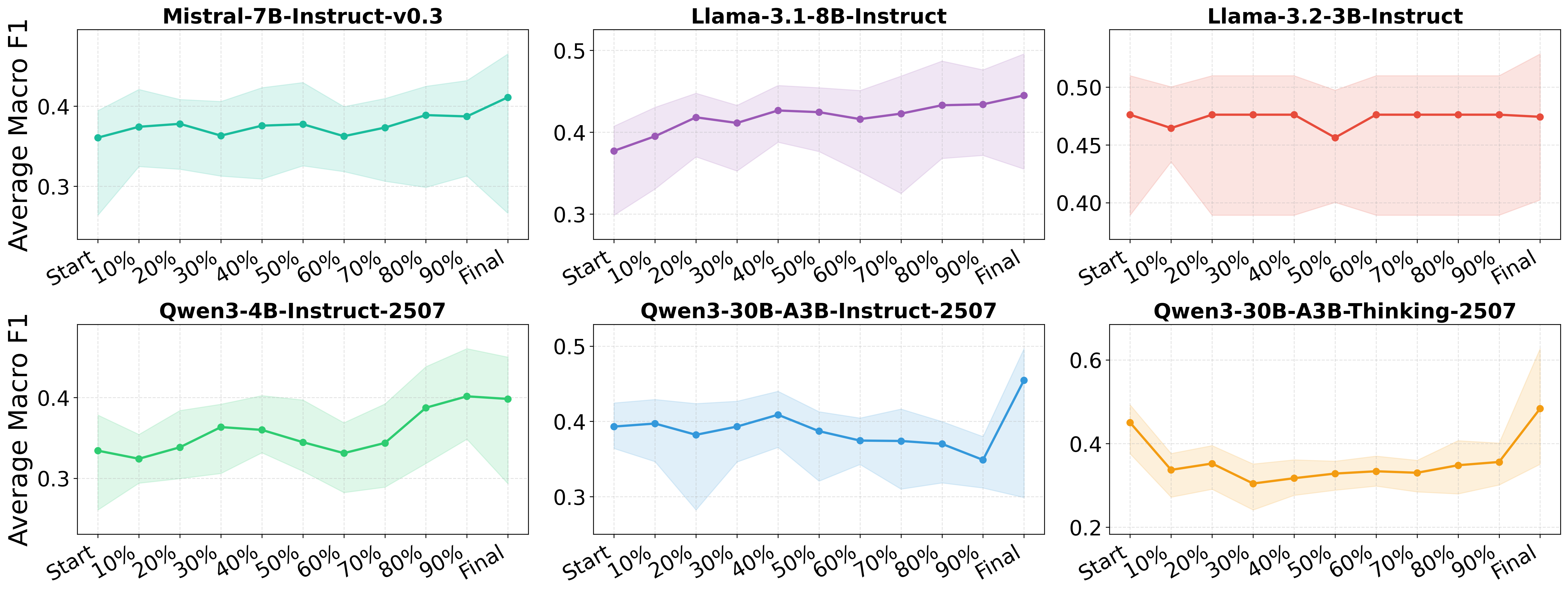}
\caption{Average macro-F1 of all layers across different token positions in the question stem.}
\label{averf1_token_position}
\vspace{-3mm}
\end{figure}

Previous sections show that position bias persists even without explicit identifiers and can be partially controlled via prompting. We now examine whether such bias is internally encoded during generation.
Focusing on Task 1 with balanced prompting as a representative setting, we probe whether models encode the correct answer position during question generation. We also report other prompting settings results in Appendix~\ref{appendix_probing}.

Following the procedure described in Section~\ref{method_2}, we extract token-level hidden states from all layers and train a lightweight MLP classifier to predict the correct answer position, evaluated using macro-F1. 
To ensure robustness, we repeat the training process five times with different random seeds and report the averaged results.


\textbf{Predictability of answer position.}  
We train MLP probes using the final-token representations, following prior work~\citep{Maar_Paperno_McDougall_Nanda_2026} demonstrating that final-token states contain informative representations for downstream prediction tasks. The hidden size is selected based on experiments reported in Appendix~\ref{probe_size}.. We report both the average F1 across layers and the peak F1 (Figure~\ref{aver_peak}). Across all models, both metrics consistently exceed random and majority baselines, indicating that \textbf{hidden representations in the question stem encode predictive signals of the correct answer position}.
\begin{wrapfigure}{r}{0.45\textwidth}
\centering
\includegraphics[width=0.45\textwidth]{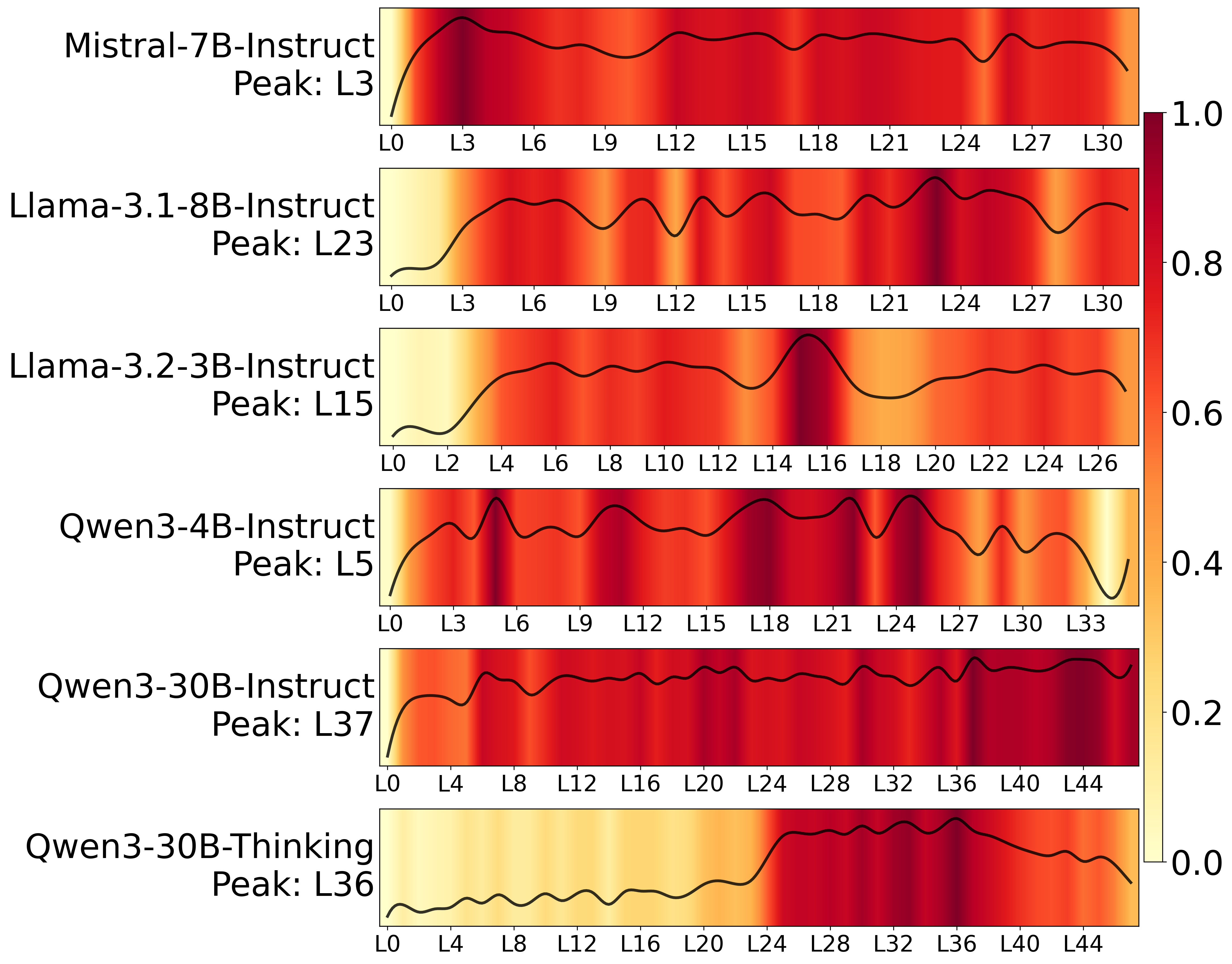}
\caption{Layer-wise probing performance for predicting correct answer positions. Colors indicate macro F1 scores, and black curves represent row-normalized relative trends across layers.}
\label{layer_wise_f1}
\vspace{-3mm}
\end{wrapfigure}

\textbf{Position-dependent encoding signals.}  
As shown in Figure~\ref{averf1_token_position}, we probe hidden states at different token positions within the question stem and examine how predictive performance varies across positions. Across most models, later tokens, particularly the final token, tend to achieve higher macro-F1 scores than earlier tokens. This suggests that answer-position information becomes more accessible in later contextualized representations, possibly because they integrate richer global contextual information.


\textbf{Layer-wise emergence of planning signals.}  
To further understand how such information is formed, we conduct layer-wise probing analysis. Results are shown in Figure~\ref{layer_wise_f1}.
We observe that predictive signals are present across a wide range of layers, rather than being confined to the final layers, suggesting that answer position information is distributed throughout the network. For example, in Qwen3-30B-Thinking, strong signals emerge around layers 24–36, while in Llama-3.2-3B-Instruct, peak performance occurs around layers 14–16.

\subsection{Intervention via Activation Steering}

Building on the probing results, we examine whether position signals causally influence correct answer position through representation-level interventions during generation. Following Section~\ref{method_3}, we apply steering vectors that shift hidden representations between target answer positions (e.g., A$\rightarrow$B) at selected layers and token positions.

\textbf{Model Setup.}
We conduct experiments on Llama-3.2-3B-Instruct, which shows strong position bias and reliable position prediction. We select 200 samples with ground-truth answer ``A'' and apply steering to bias the model toward ``B''. Steering performance is evaluated using the change rate ($A \rightarrow B$ / Baseline $A$).

\subsubsection{Intervention Results}

\begin{figure}[h]
    \centering
    \subfigure[Intervention at the final token\label{final_best_change}]{
    \includegraphics[width=0.48\textwidth]{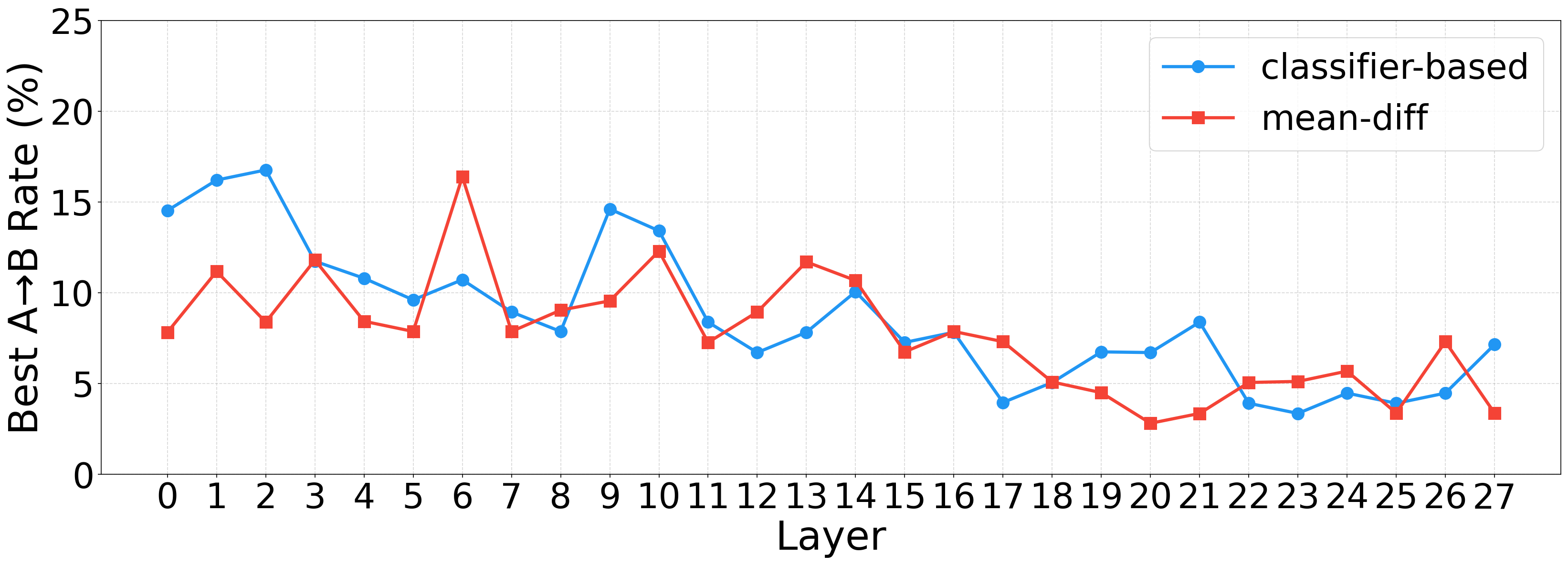}
}
\subfigure[Intervention at the penultimate token\label{penu_best_change}]{
    \includegraphics[width=0.48\textwidth]{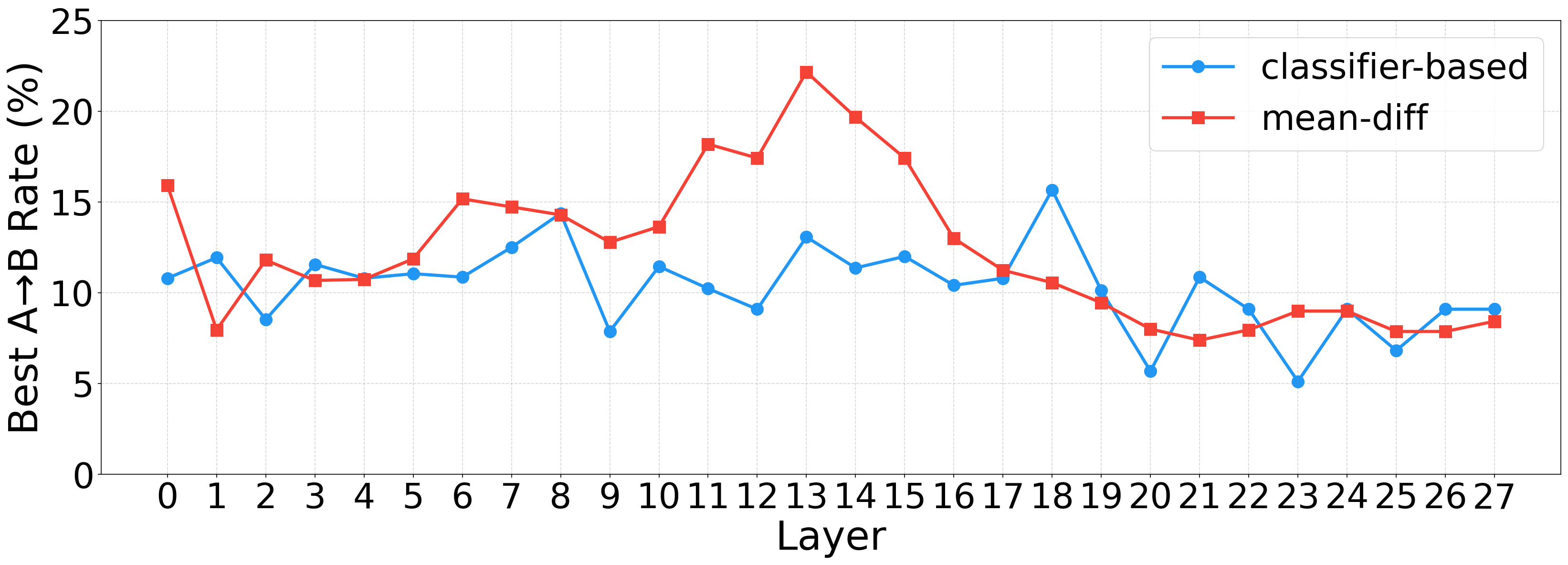}
}
    \caption{Layer-wise steering effectiveness (best A$\rightarrow$B rate) under different injection positions.
   }
    \label{fig:placeholder}
    \vspace{-3mm}
\end{figure}

\paragraph{Overall Steering Effectiveness.}
Interventions can effectively manipulate answer-position preferences under appropriate injection settings. In particular, interventions applied at the penultimate token (Figure~\ref{penu_best_change}) yield substantially stronger and more stable effects than those applied at the final token. The mean-difference method peaks in intermediate layers (11--15), reaching over 22\% A$\rightarrow$B shift at layer 13, while the classifier-based method remains comparatively stable across layers. 

\paragraph{Effect of Injection Position.}
Compared with penultimate-token interventions, steering at the final token (Figure~\ref{final_best_change}) produces weaker and less stable effects across layers. The classifier-based direction performs best in early layers (1--2), achieving around 16--17\% A$\rightarrow$B shift, but its effectiveness declines substantially in deeper layers. Meanwhile, the mean-difference method exhibits fluctuating behavior without a clear layer-wise trend.

One possible explanation is that the final token of the question stem is typically a punctuation token (e.g., ``?''), whose hidden representations exhibit limited semantic variability. As a result, steering directions constructed at this position may be less informative, reducing intervention effectiveness for both methods.

We also report layer-dependent sensitivity to the scaling factor $\alpha$ in Appendix~\ref{appendix_alpha}.

\begin{wrapfigure}{l}{0.4\textwidth}
\centering
\vspace{-5mm}
    \includegraphics[width=0.4\textwidth]{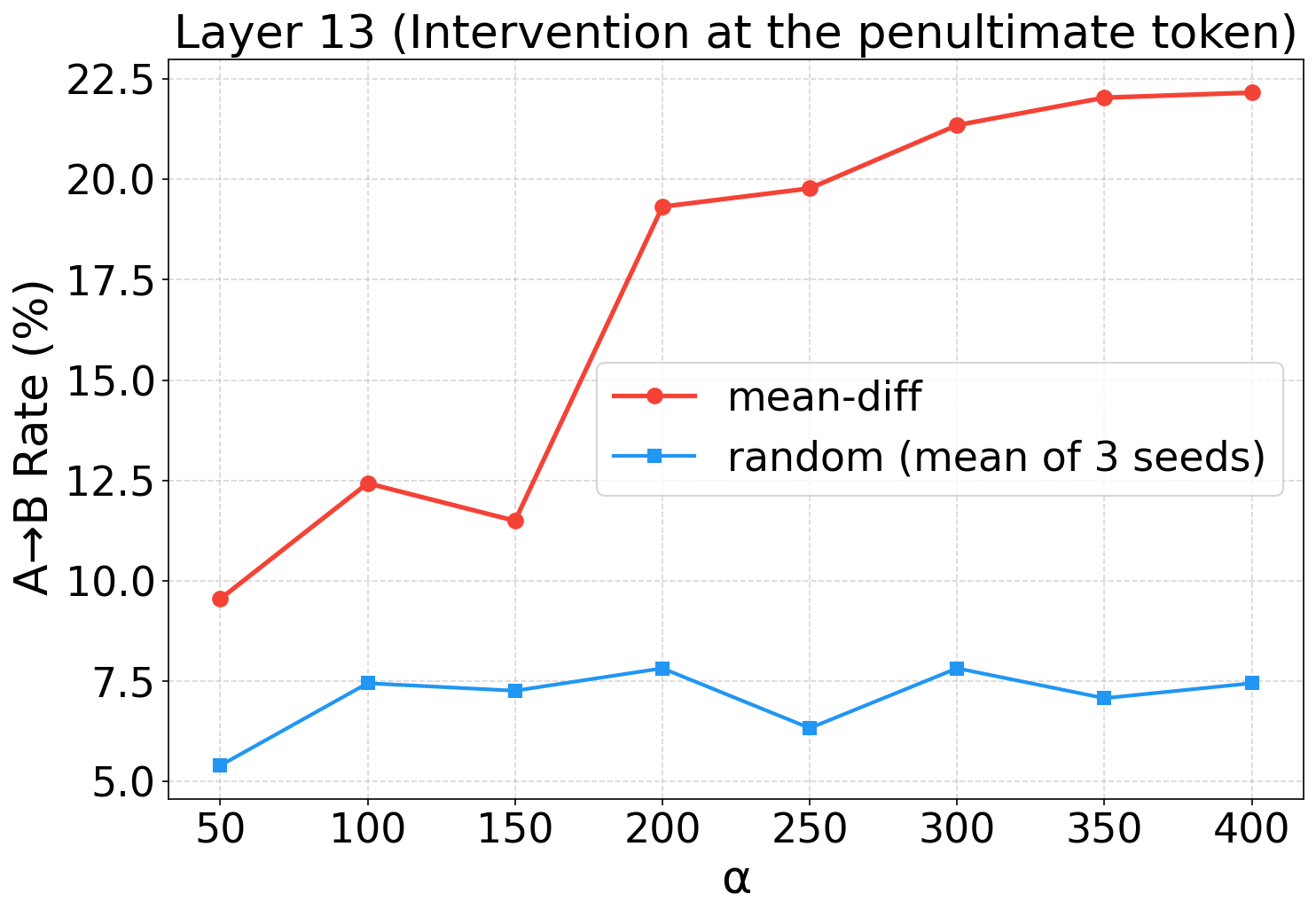}
\caption{Comparison of mean-difference and random directions at layer 13 as a function of $\alpha$.
The random baseline is computed as the average over three randomly sampled directions.}
\label{compare_direction}
\vspace{-3mm}
\end{wrapfigure}

\paragraph{Mean-Difference vs. Random Directions.}
We compare the mean-difference direction with random vectors under varying $\alpha$. Random vectors are sampled from a standard normal distribution and rescaled to match the norm of the corresponding mean-difference direction. 
As shown in Figure\ref{compare_direction}, the mean-difference method exhibits a stable, monotonic increase in the A$\rightarrow$B shift, reaching 22.2\% at $\alpha = 400$. In contrast, random vectors show high variance at small $\alpha$ due to directional instability, and plateau at around 8–10\% for larger $\alpha$, without a clear upward trend. Notably, even at $\alpha = 400$, random directions achieve only about 40\% of the effect of mean-difference.
These results suggest that the effectiveness of mean-difference steering is not driven by magnitude alone, but by the semantic alignment of the direction in representation space.

\paragraph{Target vs. Off-target Trade-off.}
\begin{wrapfigure}{r}{0.28\textwidth}
\centering
\vspace{-5mm}
\includegraphics[width=0.27\textwidth]{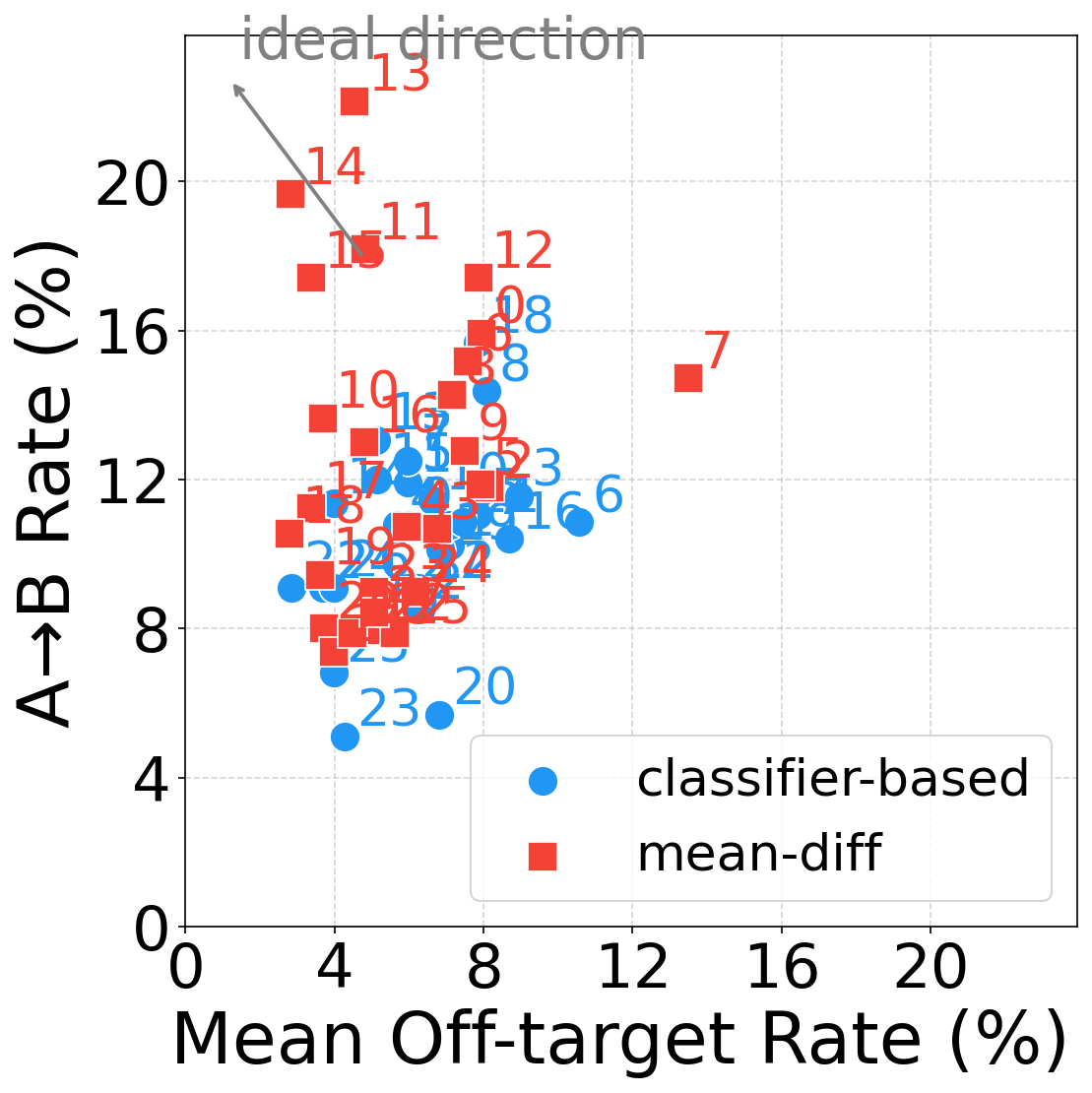}
\caption{Target vs. off-target trade-off at the penultimate token. Each point corresponds to a layer with its optimal intervention strength. ``Mean Off-target Rate'' is defined as the average rate of A$\rightarrow$C and A$\rightarrow$D.}
\label{penu_target_off}
\vspace{-3mm}
\end{wrapfigure}

We analyze the trade-off between target shifts (A$\rightarrow$B) and off-target changes (A$\rightarrow$C/D) using a 2D visualization. As shown in Figure~\ref{penu_target_off}, mean-difference steering in intermediate layers (11--14) occupies the Pareto-optimal region, achieving both high target shifts (up to 16\%) and low off-target changes (~3--6\%). In contrast, classifier-based steering remains less effective and incurs larger off-target changes, indicating less precise control. Results for steering at the final token show a similar trend and are provided in Appendix~\ref{appendix_final_token_tradeoff}.

\paragraph{Case Study}
To better understand the mechanisms behind these quantitative trends, we present representative examples in Table~\ref{case} to qualitatively analyze steering behaviors on multiple-choice questions. Across cases, \textbf{steering not only changes the predicted answer but can also modify the question semantics and option space}.

The first example is a \textbf{position control }case, although option ordering is perturbed, the semantic meaning of the question remains stable and the model preserves the correct answer. 
The second sample is a \textbf{semantic drift} case, the intervention slightly alters the question interpretation and induces a shift in the option space, while still yielding a consistent and valid answer under the new context. The third one is a \textbf{task replacement} case, which shows stronger intervention effects, where the original question is replaced by a semantically different one, leading to a different but still coherent task. 
The forth one is a \textbf{semantic refinement} case, that demonstrates mild rephrasing of the question, where the model remains robust and maintains correctness.
Overall, these examples suggest that steering can influence both answer position and question semantics.

We also observe several failure modes under intervention in Appendix~\ref{appendix_bad_case}. Steering can lead to (i) answer flips without semantic change, (ii) incoherent question–option alignment, and (iii) incorrect reasoning under modified option spaces. These results suggest that intervention directions can influence answer-position behavior, but the resulting generations do not always preserve semantic consistency or coherent reasoning. This indicates that positional steering may interact with broader generation dynamics rather than operating as an isolated control mechanism.

\begin{table}
\centering
\small
\caption{Examples illustrating different steering behaviors.}
\label{case}
\begin{tabular}{p{0.45\linewidth} p{0.45\linewidth} }
\toprule
\textbf{Baseline} & \textbf{Intervened}  \\
\midrule

\parbox[t]{\linewidth}{
\raggedright
\textbf{Q:} What is the term for the study of the interactions between organisms and their environment? \\
\textbf{Options:} \textbf{a) Ecology b) Biology }c) Environmental science d) Conservation biology \\
\textbf{Answer:} a
}
&
\parbox[t]{\linewidth}{
\raggedright
\textbf{Q:} What is the term for the study of the interactions between organisms and their environment? \\
\textbf{Options:} \textbf{a) Biology b) Ecology }c) Conservation biology d) Evolutionary biology \\
\textbf{Answer:} b
}
 \\

\midrule

\parbox[t]{\linewidth}{
\raggedright
\textbf{Q:} What is the term for the process by which cells become \textbf{specialized?} \\
\textbf{Options:} a) Differentiation b) Proliferation c) Apoptosis d) Transdifferentiation \\
\textbf{Answer:} a
}
&
\parbox[t]{\linewidth}{
\raggedright
\textbf{Q:} What is the term for the process by which cells become \textbf{divide and form new tissues?} \\
\textbf{Options:} a) Differentiation b) Proliferation c) Apoptosis d) Senescence \\
\textbf{Answer:} b
}
 \\

\midrule

\parbox[t]{\linewidth}{
\raggedright
\textbf{Q:} What is the world's largest \textbf{desert?} \\
\textbf{Options:} a) Sahara Desert b) Gobi Desert c) Mojave Desert d) Atacama Desert \\
\textbf{Answer:} a
}
&
\parbox[t]{\linewidth}{
\raggedright
\textbf{Q:} What is the largest \textbf{continent?} \\
\textbf{Options:} a) Africa b) Asia c) North America d) South America \\
\textbf{Answer:} b
}
\\

\midrule

\parbox[t]{\linewidth}{
\raggedright
\textbf{Q:} What is the term for the movement of people from one city to \textbf{another?} \\
\textbf{Options:} a) Migration b) Immigration c) Emigration d) Internal migration \\
\textbf{Answer:} a
}
&
\parbox[t]{\linewidth}{
\raggedright
\textbf{Q:} What is the term for the movement of people from one city to \textbf{another cities?} \\
\textbf{Options:} a) Intra-urban migration b) Inter-urban migration c) Suburbanization d) Urbanization \\
\textbf{Answer:} b
}
\\

\bottomrule
\end{tabular}
\vspace{-3mm}
\end{table}

\section{Conclusion}
\label{conclusion}
In this work, we conduct a systematic study of position bias in multiple-choice question generation. Across a diverse set of LLMs, we show that answer positions are not uniformly distributed, but exhibit consistent and model-specific preferences. Through controlled prompting and probing analyses, we further demonstrate that hidden representations in the question stem contain predictive signals of the eventual answer position, with later contextualized representations often encoding stronger position-related information.

To better understand the underlying mechanism, we investigate whether answer-position behavior can be manipulated through representation-level interventions. Our results show that steering directions derived from hidden-state differences can significantly shift answer distributions, indicating that positional preferences are at least partially encoded in the model’s internal representations. At the same time, intervention experiments and failure cases reveal that modifying positional behavior may also affect semantic coherence and generation dynamics, suggesting that position bias is intertwined with broader generation processes rather than operating as an isolated mechanism.


\textbf{Limitations.}
First, although probing and intervention results suggest that answer-position information is encoded in hidden representations, they do not provide definitive evidence of explicit symbolic planning. Instead, our findings are better interpreted as evidence of implicit planning-like mechanisms reflected in model representations and causal interventions.
Second, intervention effectiveness depends strongly on injection position and representation quality. In particular, steering at the final token of the question stem often yields weaker effects, possibly because the final token is frequently punctuation (e.g., ``?’’), whose representations contain limited semantic variability.
Third, our experiments focus on controlled MCQ generation settings with fixed answer formats and prompting templates, which may not fully capture open-ended or real-world scenarios.
Finally, our analysis is limited to a finite set of LLMs and VLMs under deterministic decoding settings. Positional preferences may vary under different decoding strategies, prompting styles, or future model architectures.

\textbf{Broader Impact.} This work improves the understanding of positional biases in LLMs and provides tools for analyzing and intervening in internal representations. However, techniques for steering model behavior may also be misused to manipulate model outputs or amplify undesirable biases. We encourage future work on developing safer and more transparent intervention methods.

\bibliographystyle{plainnat}
\bibliography{cite}

\appendix

\section{Details of Steering Vector Construction}
\label{appendix:steering_vectors}
\paragraph{Mean Difference.}

For a given layer $l$ and token position $i$ in the question stem, we group hidden representations according to their predicted answer positions:

\begin{equation} 
\mathcal{H}_A^{(l,i)} = \{ H_{x}^{(l,i)} \mid \text{Answer}(H_{x}^{(l,i)}) = src \}, \quad
\mathcal{H}_D^{(l,i)} = \{ H_{x}^{(l,i)} \mid \text{Answer}(H_{x}^{(l,i)}) = tgt \}.
\end{equation}

We then compute the centroid (mean representation) of each group:

\begin{equation} 
\mu_{src}^{(l,i)} = \frac{1}{|\mathcal{H}_{src}^{(l,i)}|} \sum_{H \in \mathcal{H}_{src}^{(l,i)}} H, \quad
\mu_{tgt}^{(l,i)} = \frac{1}{|\mathcal{H}_{tgt}^{(l,i)}|} \sum_{H \in \mathcal{H}_{tgt}^{(l,i)}} H.
\end{equation}

The steering vector is defined as the difference between the two centroids:

\begin{equation}
v_{src \rightarrow tgt}^{(l,i)} = \mu_{tgt}^{(l,i)} - \mu_{src}^{(l,i)}.
\end{equation}

This vector represents the average direction in representation space that shifts the model’s internal state from favoring option $src$ to favoring option $tgt$. 
Intuitively, it captures the linear feature associated with positional preference encoded in the hidden states.

\label{classifier_based_vector}
\paragraph{Classifier-Based Steering.}
In addition to the mean-difference approach, we construct steering vectors using a linear probe trained to distinguish between source and target answer positions.

Specifically, for each layer $l$ and token position $i$, we collect hidden representations $H_x^{(l,i)}$ and assign binary labels according to the answer position, where the source class (e.g., $A$) is labeled as $0$ and the target class (e.g., $B$) as $1$. 
We then train an $\ell_2$-regularized logistic regression classifier:

\begin{equation}
p(y=1 \mid H_x^{(l,i)}) = \sigma(w^{(l,i)} \cdot H_x^{(l,i)} + b),
\end{equation}

where $w^{(l,i)}$ and $b$ denote the learned weight and bias, and $\sigma(\cdot)$ is the sigmoid function. 
The classifier is trained with an $\ell_2$-regularized logistic objective, where the regularization term is scaled by the hyperparameter $C$ (inverse regularization strength in sklearn convention).

We take the learned weight vector as the steering direction:
\begin{equation}
v_{src \rightarrow tgt}^{(l,i)} = w^{(l,i)}.
\end{equation}

\paragraph{Sign Alignment.}
Since the logistic regression solution is invariant to a global sign flip, we explicitly align the direction of $w^{(l,i)}$ to ensure consistent intervention semantics. 
Let $\mu_{\text{src}}^{(l,i)}$ and $\mu_{\text{tgt}}^{(l,i)}$ denote the mean representations of the source and target classes, respectively. 
We enforce:

\begin{equation}
w^{(l,i)} \cdot (\mu_{\text{tgt}}^{(l,i)} - \mu_{\text{src}}^{(l,i)}) > 0.
\end{equation}

If this condition is not satisfied, we multiply both $w^{(l,i)}$ and $b$ by $-1$. 
After alignment, adding the steering vector (i.e., $+\alpha v$) consistently shifts representations toward the target class.



\section{Experimental Setup}
\label{exp1_setup}

During data collection with LLMs, we use deterministic decoding by setting temperature to 0 and disabling sampling. All open-source models are deployed on NVIDIA A100 GPUs.

We use the following API model mappings in our experiments:

\begin{itemize}

    \item \texttt{deepseek-chat} $\rightarrow$ DeepSeek-V3-0324

    \item \texttt{deepseek-reasoner} $\rightarrow$ DeepSeek-R1-0528

    \item \texttt{gemini-3-pro} $\rightarrow$ Gemini 3 Pro (preview version)

\end{itemize}

\section{Bias Evaluation Metrics}
\label{bias_m}
\textbf{Metrics}: we use two metrics to assess the LLMs correct answer position bias. The first is Bias Score~\citet{Reif_Schwartz_2024} as follows.

Let $f_1, f_2, \dots, f_K$ denote the frequencies of $K$ categories.
We first normalize them into probabilities:

\begin{equation}
p_i = \frac{f_i}{\sum_{j=1}^{K} f_j}
\end{equation}

\begin{equation}
\text{BSD} =
\sum_{i=1}^{K} \left|p_i - \frac{1}{K}\right|
\end{equation}

The second metric is standard deviation of recalls
(RStd) ~\citet{Zheng_Zhou_Meng_Zhou_Huang_2024}.

\begin{equation}
\text{RStd}= \frac{\sqrt{\frac{1}{K}\sum_{i=1}^{K}(p_i - \frac{1}{K})^2}}{\frac{1}{K}}
\end{equation}
Intuitively, RStd is low when model performance is
similar on all classes, and high when it performs well on some classes but poorly on others.
When we set the number of option $K=4$, the largest BSD is 1.5 and the largest RSD is 1.732.

\textbf{Chi-square test}. 

In addition, we perform a chi-square goodness-of-fit test to examine whether the observed answer distribution significantly deviates from a uniform distribution:

\begin{equation}
\chi^2 = \sum_{i=1}^{K} \frac{(f_i - E)^2}{E}, \quad \text{where } E = \frac{1}{K} \sum_{j=1}^{K} f_j
\end{equation}

A significant test result indicates that the model's answer distribution is unlikely to be uniform, providing statistical evidence of position bias. We report the $\chi^2$ statistic together with the corresponding $p$-value.

\section{Additional Position Bias Results}
\label{metrics_results}
\begin{table}
\centering
\small
\caption{Task 1 (K-MCQ): Position bias in knowledge-based MCQs. }
\label{task1-1}
\begin{tabular}{lccc}
\hline
\Xhline{1.2pt}
Model & BSD $\downarrow$ & RStd $\downarrow$  & $\chi^2$ (p-value)\\
\hline
Mistral-7B-Instruct-v0.3&0.23&0.26 &51.70(p < 0.001)\\
Llama-3.1-8B-Instruct &0.45 &0.53 & 283.38(p < 0.001)\\
Llama-3.2-3B-Instruct &0.44&0.54 &294.11(p < 0.001)\\
Qwen3-4B-Instruct-2507 &0.55&0.63 &390.94(p < 0.001)\\
Qwen3-30B-Instruct-2507 &0.55 &0.70 & 470.46(p < 0.001)\\
Qwen-30B-Thinking-2507 &0.42 &0.52& 260.08(p < 0.001)\\
DeepSeek-Reasoner & 0.49 &  0.61&373.33(p < 0.001)\\
DeepSeek-Chat & 0.56 & 0.66 &439.37(p < 0.001)\\
Gemini-2.5-Flash & 0.69& 0.70&494.50(p < 0.001)\\
Gemini-3-pro & 0.66&0.71&507.93(p < 0.001)\\
\hline
\Xhline{1.2pt}
\end{tabular}
\end{table}

\begin{table}
\centering
\small
\caption{Task 1: Bias metrics across five disciplines.}
\label{task1_result_disc}
\setlength{\tabcolsep}{0.3mm}
\begin{tabular}{c|cc|cc|cc|cc|cc}
\hline
\Xhline{1.2pt}
\multirow{3}{*}{\textbf{Models}} 
& \multicolumn{10}{c}{Disciplines} \\
\cline{2-11}

& \multicolumn{2}{c|}{Mathematics}
& \multicolumn{2}{c|}{Chemistry}
& \multicolumn{2}{c|}{Biology}
& \multicolumn{2}{c|}{History}
& \multicolumn{2}{c}{Geography} \\

& BSD $\downarrow$& RStd$\downarrow$
& BSD $\downarrow$& RStd$\downarrow$
& BSD $\downarrow$& RStd$\downarrow$
& BSD $\downarrow$& RStd$\downarrow$
& BSD $\downarrow$& RStd$\downarrow$\\
\hline

Mistral-7B-Instruct-v0.3 
 
& 0.28 & 0.34
& 0.14 & 0.18 
& 0.23 & 0.29 
& 0.24 & 0.30
& 0.20 & 0.24 \\

Llama-3.1-8B-Instruct 
 
& 0.64 & 0.77 
& 0.46 & 0.51
& 0.34 & 0.37
& 0.64 & 0.74
& 0.52 & 0.55 \\

Llama-3.2-3B-Instruct 

& 0.66 & 0.76
& 0.22 & 0.27
& 0.19&  0.21
& 1.11 & 1.29
& 0.67 & 0.77 \\

Qwen3-4B-Instruct-2507 

& \textbf{0.72} & \textbf{0.86}
& 0.54 & 0.69 
& 0.51 & 0.67 
& 0.67 & 0.81
& \textbf{0.66} & \textbf{0.74} \\

Qwen3-30B-A3B-Instruct-2507 

& 0.55 & 0.59
& 0.74 & 0.77
& \textbf{0.62} & \textbf{0.69}
& \textbf{0.64} & \textbf{0.81}
& 0.93 & 1.08 \\

Qwen3-30B-A3B-Thinking-2507 

& 0.78 & 0.94
& 0.49 & 0.54
& 0.44 & 0.52
& 0.59 & 0.65 
& 0.50 & 0.66\\
Deepseek-chat 

& 0.47 & 0.52 
& 0.49 &  0.55
& 0.75 &0.76
& 0.74 & 0.87 
& 0.70& 0.84\\
Deepseek-reasoner
 
& 0.55& 0.62 
& 0.54 & 0.65 
& 0.60 & 0.63
& 0.66 & 0.82
& 0.62 & 0.67 \\
Gemini-2.5-flash
 
& 0.41& 0.52 
&\textbf{0.87 }& \textbf{0.89}  
& 0.80 & 0.86
& 0.90 & 0.90
& 0.82& 0.83\\
Gemini-3-pro

&0.43&  0.55 
&0.73  & 0.77
& 0.77 & 0.79
& 0.71 &0.77
& 0.64&0.70  \\
\hline
\Xhline{1.2pt}
\end{tabular}
\end{table}

\begin{table}
\centering
\small
\caption{Task 1: Chi-square statistics ($\chi^2$) with corresponding $p$-values across five disciplines.}
\label{task1_disc_chi}
\setlength{\tabcolsep}{0.1mm}
\begin{tabular}{c|c|c|c|c|c}
\hline
\Xhline{1.2pt}
\multirow{2}{*}{\textbf{Models}} 
& \multicolumn{5}{c}{Disciplines} \\
\cline{2-6}

& \multicolumn{1}{c|}{Mathematics}
& \multicolumn{1}{c|}{Chemistry}
& \multicolumn{1}{c|}{Biology}
& \multicolumn{1}{c|}{History}
& \multicolumn{1}{c}{Geography} \\

\hline

Mistral-7B-Instruct-v0.3 
& 36.47 (6.0e-08) & 2.63 (0.453) & 7.11 (0.069) & 33.32 (2.8e-07) & 14.09 (0.0028) \\

Llama-3.1-8B-Instruct 
& 118.96 (1.3e-25) & 52.60 (2.2e-11) & 28.12 (3.4e-06) & 110.60 (8.2e-24) & 60.84 (3.9e-13) \\

Llama-3.2-3B-Instruct 
& 116.92 (3.6e-25) & 14.60 (0.0022) & 8.52 (0.036) & 329.02 (5.2e-71) & 119.96 (7.9e-26) \\

Qwen3-4B-Instruct-2507 
& 146.92 (1.2e-31) & 93.72 (3.5e-20) & 90.28 (1.9e-19) & 132.12 (1.9e-28) & 110.48 (8.7e-24) \\

Qwen3-30B-A3B-Instruct-2507 
& 65.46 (4.0e-14) & 112.88 (2.6e-24) & 90.53 (1.7e-19) & 123.87 (1.1e-26) & 220.84 (1.3e-47) \\

Qwen3-30B-A3B-Thinking-2507 
& 141.65 (1.7e-30) & 59.32 (8.2e-13) & 55.12 (6.5e-12) & 85.48 (2.0e-18) & 86.28 (1.4e-18) \\

Deepseek-chat 
& 55.00 (6.9e-12) & 60.60 (4.4e-13) & 116.92 (3.6e-25) & 153.04 (5.8e-33) & 139.27 (5.4e-30) \\

Deepseek-reasoner
& 74.20 (5.4e-16) & 84.56 (3.2e-18) & 80.92 (1.9e-17) & 135.72 (3.2e-29) & 89.40 (2.9e-19) \\

Gemini-2.5-flash
& 55.08 (6.6e-12) & 157.72 (5.7e-34) & 147.36 (9.8e-32) & 163.48 (3.3e-35) & 137.76 (1.1e-29) \\

Gemini-3-pro
& 61.56 (2.7e-13) & 119.80 (8.5e-26) & 127.48 (1.9e-27) & 119.32 (1.1e-25) & 99.36 (2.1e-21) \\

\hline
\Xhline{1.2pt}
\end{tabular}
\end{table}

\begin{figure}[h]
    \vspace{-3mm}
    \centering
    \includegraphics[width=0.9\linewidth]{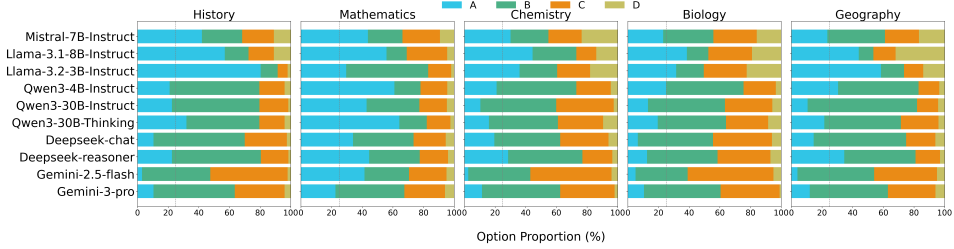}
    \caption{Distribution of correct answer positions in K-MCQ generated by 10 LLMs across five disciplines. }
    \label{task1_figure}
    \vspace{-3mm}
\end{figure}

\paragraph{Results for Task 1 (Knowledge-based MCQ Generation).}
As shown in Table~\ref{task1-1}, all models exhibit non-negligible position bias, as indicated by consistently non-zero BSD and RStd values, suggesting that correct answers are not uniformly distributed across options even in generation settings.

We observe clear variation across model families. Smaller or instruction-tuned models (e.g., Mistral-7B, Llama-3.1-8B) generally show lower bias, while larger models (e.g., Qwen3-30B-A3B-Instruct) exhibit substantially stronger bias. This indicates that scaling does not necessarily mitigate position bias.

We also report metrics and Chi-square test on each discipline in Table~\ref{task1_result_disc} and ~\ref{task1_disc_chi}. Across disciplines, History and Geography tend to induce stronger bias, whereas Mathematics shows relatively lower bias. Notably, we plot the option proportion distribution across five disciplines for each model (Figure~\ref{task1_figure}). Most models exhibit a clear tendency to place correct answers in option ``A’’ in Mathematics.

Finally, reasoning-oriented variants show reduced bias compared to their base counterparts, suggesting that alignment or reasoning mechanisms may help alleviate position bias.

Overall, position bias is pervasive, varies across models and domains, and is not trivially resolved by increasing model scale.

\begin{table}
\vspace{-5mm}
\centering
\small
\caption{Task 2 (RC-MCQ): position bias in reading comprehension tasks..}
\label{task2}
\begin{tabular}{lccc}
\hline
\Xhline{1.2pt}
Model & BSD $\downarrow$ & RStd $\downarrow$   & $\chi^2$ (p-value)\\
\hline
Mistral-7B-Instruct-v0.3&0.20 &0.25& 23.83 (p<0.001)\\
Llama-3.1-8B-Instruct & 0.41&0.43 &89.61 (p<0.001)\\
Llama--3.2-3B-Instruct &0.65 &0.35&61.04 (p<0.001)\\
Qwen3-4B-Instruct-2507 &0.65&0.75&270.12 (p<0.001)\\
Qwen3-30B-Instruct-2507 &0.89 &0.94 &428.21 (p<0.001)\\
Qwen-30B-Thinking-2507 & 0.75&  0.76&284.13 (p<0.001)\\
DeepSeek-Reasoner & 0.85 & 1.03& 531.88 (p<0.001) \\
DeepSeek-Chat & 0.76 & 0.77 & 299.11 (p<0.001)\\
Gemini-2.5-Flash &  0.85 & 0.85& 365.17 (p<0.001)\\
Gemini-3-pro & 0.57 & 0.58 &171.93 (p<0.001) \\
\hline
\Xhline{1.2pt}
\end{tabular}
\end{table}
\paragraph{Results for Task 2 (Reading Comprehension MCQ Generation).}
As shown in Table~\ref{task2}, we further evaluate position bias under a comprehension-based question-answer generation setting. Compared to Task 1, we observe that position bias remains consistently present across all models, with BSD and RStd values generally at a comparable or even higher level.

In particular, larger models such as Qwen3-30B-Instruct-2507, DeepSeek-Reasoner and Gemini-2.5-Flash exhibit relatively strong bias (BSD > 0.85), indicating that incorporating a reading comprehension process does not mitigate position bias. Instead, the bias appears to persist even when models are required to ground their answers in a given context.

Interestingly, reasoning-oriented variants such as Qwen-30B-Thinking-2507 show slightly reduced bias compared to their base counterparts, suggesting that enhanced reasoning or alignment mechanisms may partially regularize answer position distributions. However, this effect is limited, as the overall bias remains substantial across models.


\begin{table}
\vspace{-5mm}
\centering
\small
\caption{Task 3 (VL-MCQ): position bias in multimodal MCQ generation.}
\label{task3}
\begin{tabular}{lccc}
\hline
\Xhline{1.2pt}
Model & BSD$\downarrow$ & RStd$\downarrow$  & $\chi^2$ (p-value)\\
\midrule
Llama-3.2-11B-Vision-Instruct & 0.65& 0.79  &303.08 (p < 0.001)\\
Llava-v1.6-mistral-7b-hf
 & \textbf{0.83}& \textbf{0.97} & 437.06(p < 0.001)\\
Qwen3-VL-235B-Instruct & 0.82 &  0.83 &330.71(p < 0.001)\\
Qwen3-VL-235B-Thinking & 0.49 & 0.67 &219.18(p < 0.001) \\
Gemini-3-pro-image & 0.55 & 0.55 &141.87(p < 0.001)\\
\hline
\Xhline{1.2pt}
\end{tabular}
\vspace{-5mm}
\end{table}

\paragraph{Results for Task 3 (Vision-Language MCQ Generation).}
We further extend our analysis to a multimodal setting, as shown in Table~\ref{task3}. Despite the inclusion of visual inputs, all evaluated VLMs still exhibit noticeable position bias.

Compared to text-only settings, the magnitude of bias remains substantial. For instance, LLaVA-v1.6-Mistral-7B exhibits the strongest bias (BSD = 0.83, RSTD = 0.97), while Qwen3-VL-235B-Thinking and Gemini-3-Pro-Image-Preview achieve comparatively lower values. This trend again suggests that reasoning-enhanced variants can partially mitigate, but not eliminate, position bias.


\section{Additional Results for Identifier-Free Setting}
\label{Appendix_ID}
\begin{figure}[h]
\centering

\subfigure[Task 2: BSD (with vs. w/o ID)]{
    \includegraphics[width=0.45\textwidth]{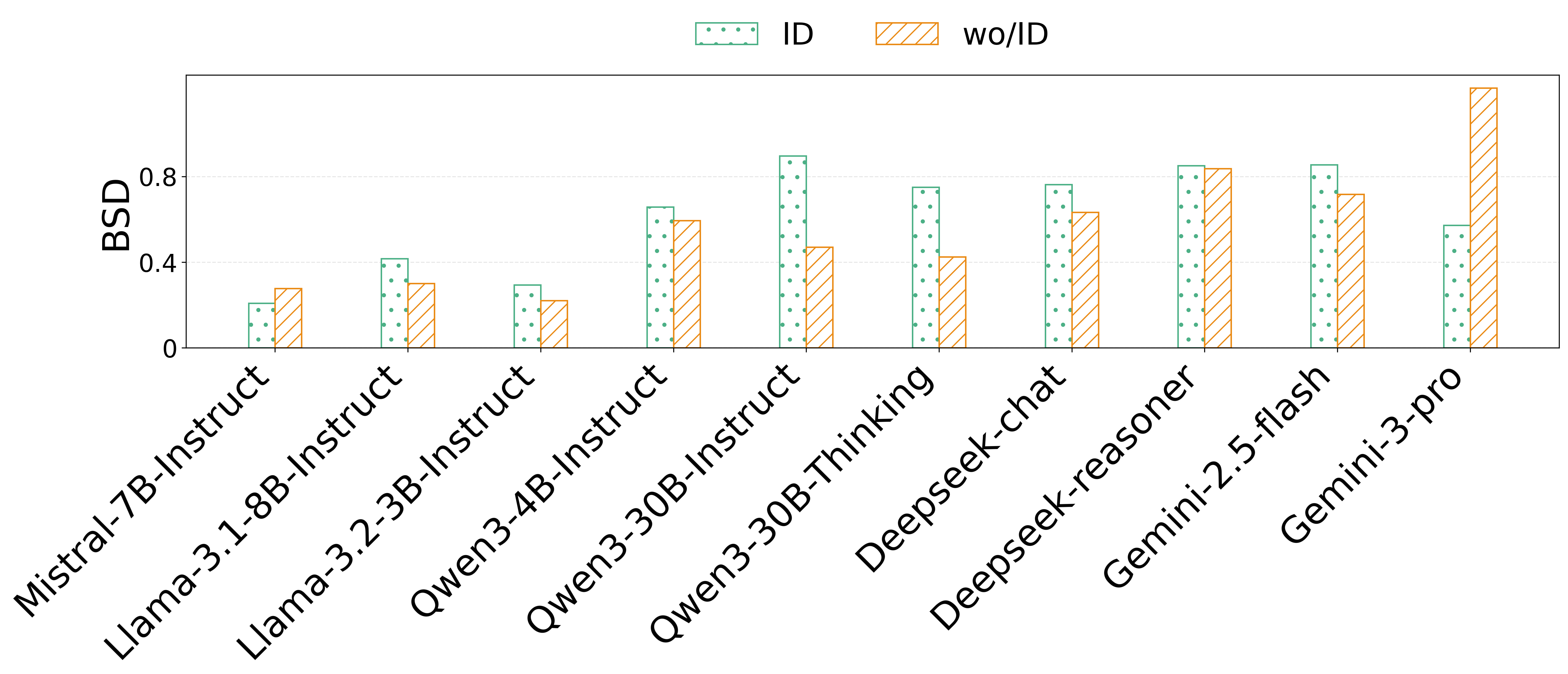}
}
\subfigure[Task 2: Distribution (with vs. w/o ID)]{
    \includegraphics[width=0.45\textwidth]{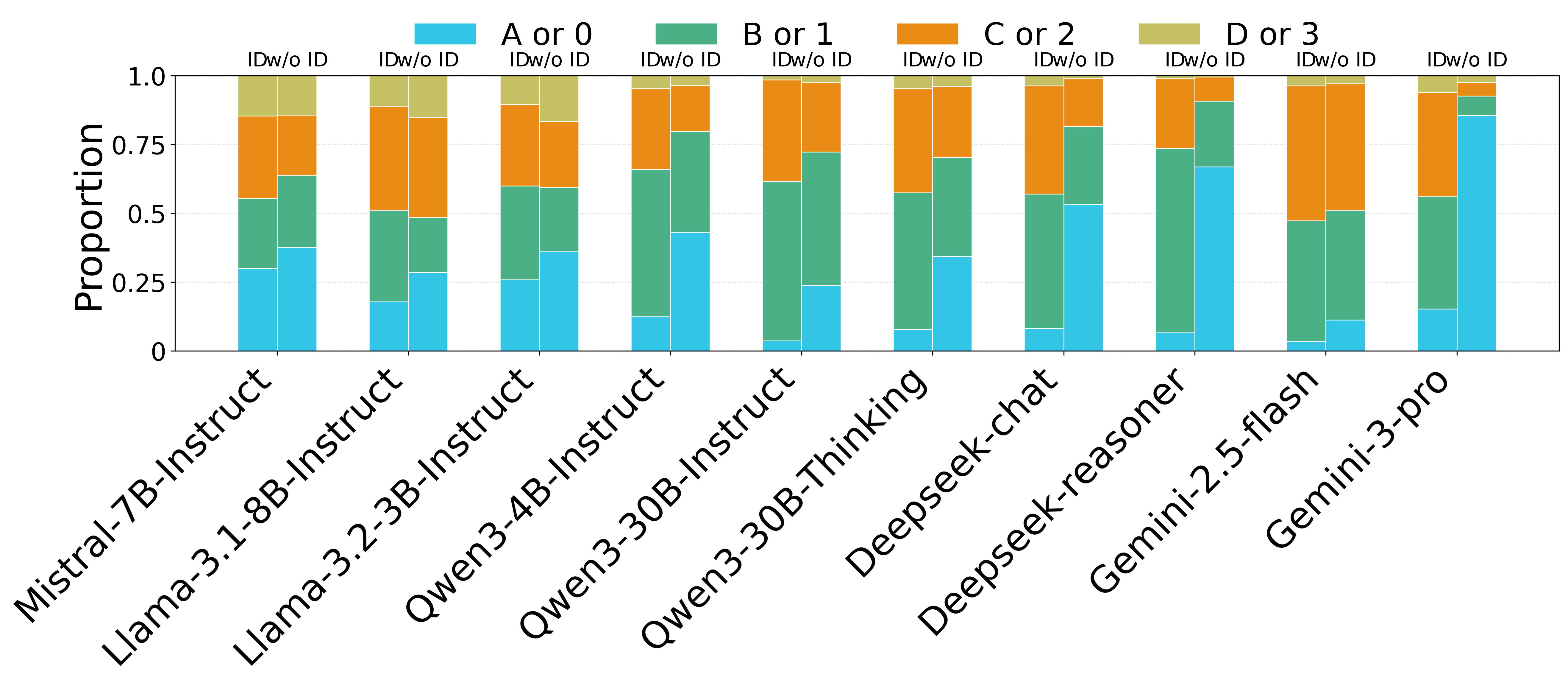}
}

\subfigure[Task 3: BSD (with vs. w/o ID)]{
    \includegraphics[width=0.45\textwidth]{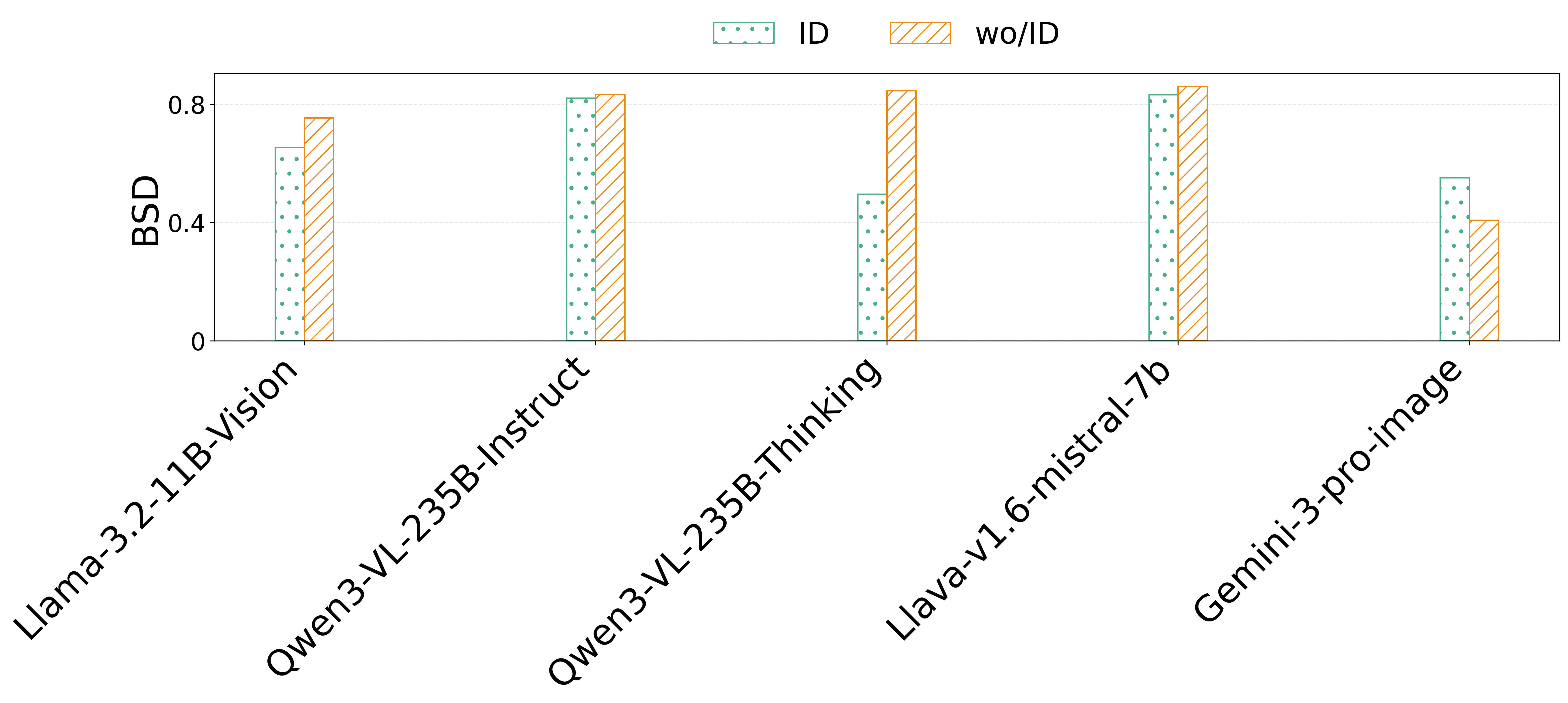}
}
\subfigure[Task 3: Distribution (with vs. w/o ID)]{
    \includegraphics[width=0.45\textwidth]{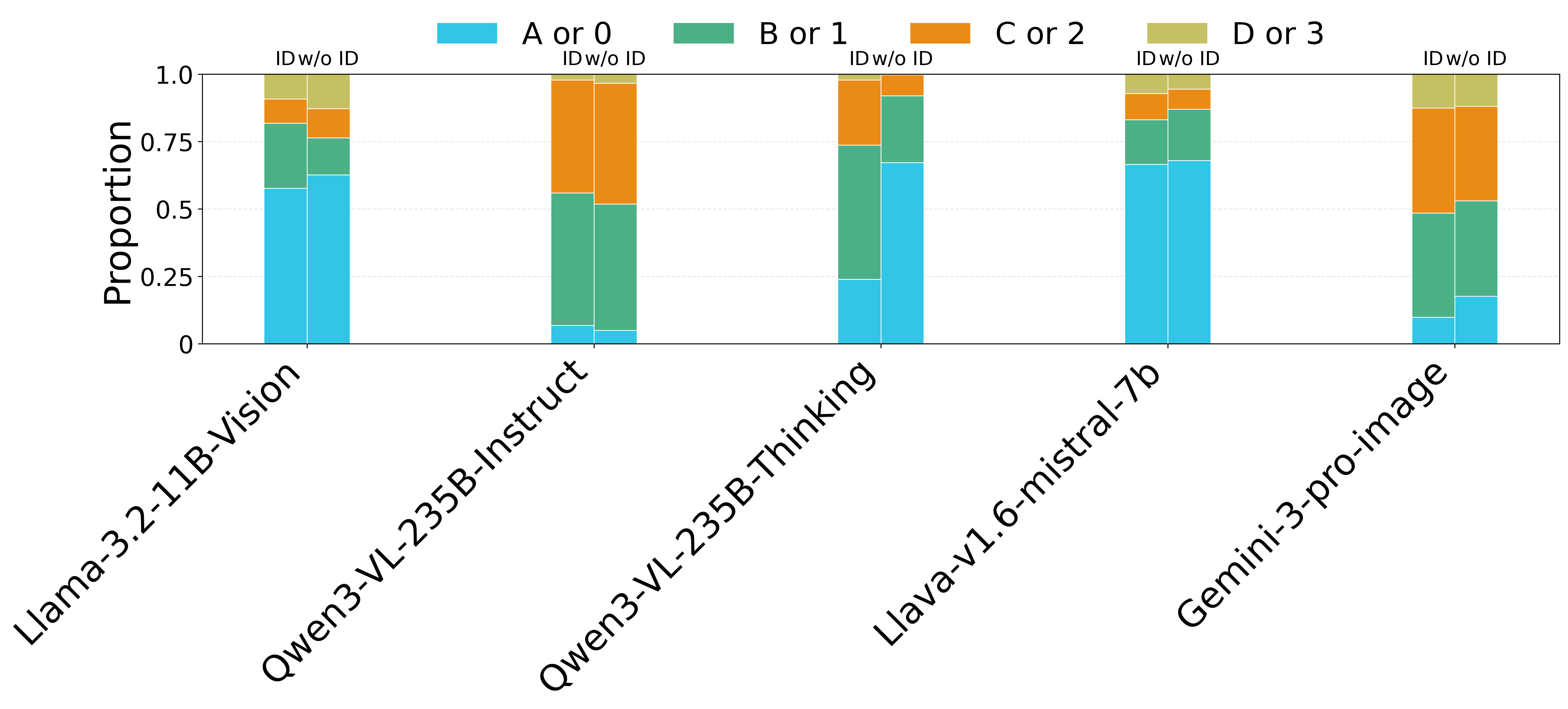}
}

\caption{
Comparison of position bias with and without option identifiers on Task 2 and 3. 
While BSD changes vary across models, a consistent shift toward the first position is observed in the identifier-free setting, indicating intrinsic ordering preferences.
}
\label{Task23_normal_ID_woID}
\end{figure}

We present the results for Task 2 and Task 3 in Figure~\ref{Task23_normal_ID_woID}. 

Consistent with Task 1, removing option identifiers leads to a strong preference for the first generated position across models.

\section{Statistical Analysis of Identifier Removal}
\label{Appendix_ID_stats}

To assess whether removing option identifiers systematically increases first-position preference, we fit a mixed-effects logistic regression model:

\begin{equation}
\texttt{selected\_first} \sim \texttt{condition} + (1|\texttt{question}) + (1|\texttt{model})
\end{equation}

where \texttt{selected\_first} is a binary variable indicating whether the correct answer appears in the first position, and \texttt{condition} denotes the labeled versus unlabeled setting. Random intercepts for both question and model are included to account for instance-level and model-level variability. In addition, we perform one-sided Fisher's exact tests for each individual model ($H_1$: the unlabeled condition increases first-position preference).

The mixed-effects model reveals a significant positive effect of the unlabeled condition on first-position preference ($\beta = 0.565$, $SE = 0.032$, $z = 17.66$, $p < 0.001$). The corresponding odds ratio is 1.76 (95\% CI [1.653, 1.874]), indicating that removing option identifiers increases the odds of generating a first-position correct answer by approximately 76\%. These findings suggest that option identifiers provide structured positional cues that mitigate primacy bias during MCQ generation.

\begin{table}[t]
\centering
\small
\caption{Model-wise comparison of first-position preference under labeled and unlabeled settings. Odds ratios (OR) and one-sided Fisher exact tests evaluate whether removing option identifiers significantly increases first-position preference. *** indicates $p < .001$; n.s. indicates $p \geq .05$.}
\label{tab:fisher_identifier}
\begin{tabular}{lcccl}
\toprule
\textbf{Model} & \textbf{Unlabeled (\%)} & \textbf{Labeled (\%)} & \textbf{OR} & \textbf{$p$-value} \\
\midrule
Mistral-7B-Instruct-v0.3 & 41.2 & 33.1 & 1.416 & $4.20\times10^{-4}$*** \\
Llama-3.1-8B-Instruct & 56.8 & 47.9 & 1.429 & $4.28\times10^{-5}$*** \\
Llama-3.2-3B-Instruct & 57.9 & 47.1 & 1.542 & $9.06\times10^{-7}$*** \\
Qwen3-4B-Instruct-2507 & 40.8 & 31.7 & 1.485 & $1.39\times10^{-5}$*** \\
Qwen3-30B-Instruct-2507 & 42.3 & 19.9 & 2.956 & $4.13\times10^{-27}$*** \\
Qwen3-30B-Thinking-2507 & 42.2 & 29.3 & 1.761 & $1.65\times10^{-9}$*** \\
DeepSeek-Reasoner & 55.2 & 28.5 & 3.091 & $3.23\times10^{-34}$*** \\
DeepSeek-Chat & 22.9 & 17.0 & 1.448 & $6.01\times10^{-4}$*** \\

Gemini-2.5-flash & 10.8 & 11.2 & 0.964 & .628 (n.s.) \\
Gemini-3-pro & 23.9 & 13.4 & 2.030 & $9.73\times10^{-10}$*** \\

\bottomrule
\end{tabular}
\end{table}

As shown in Table~\ref{tab:fisher_identifier}, at the individual-model level, 9 out of 10 evaluated models show significantly stronger first-position preference under the identifier-free setting (all $p < 0.001$), with odds ratios ranging from 1.42 (Llama-3.1-8B) to 3.09 (DeepSeek-Reasoner). This result indicates that primacy bias broadly and consistently emerges when explicit option identifiers are removed. The only exception is Gemini-2.5-Flash, which exhibits no significant difference between the two conditions ($p = 0.628$). Interestingly, this model instead shows a tendency toward later positions under the unlabeled setting, with the third position becoming the most frequent (48.5\%), suggesting architectural heterogeneity in how positional bias manifests across models.

\section{Additional Results for Balanced Prompting}
\label{Appendix_balance}
\begin{figure}
\centering
\subfigure[Task 2: BSD (standard vs. balanced)]{
    \includegraphics[width=0.45\textwidth]{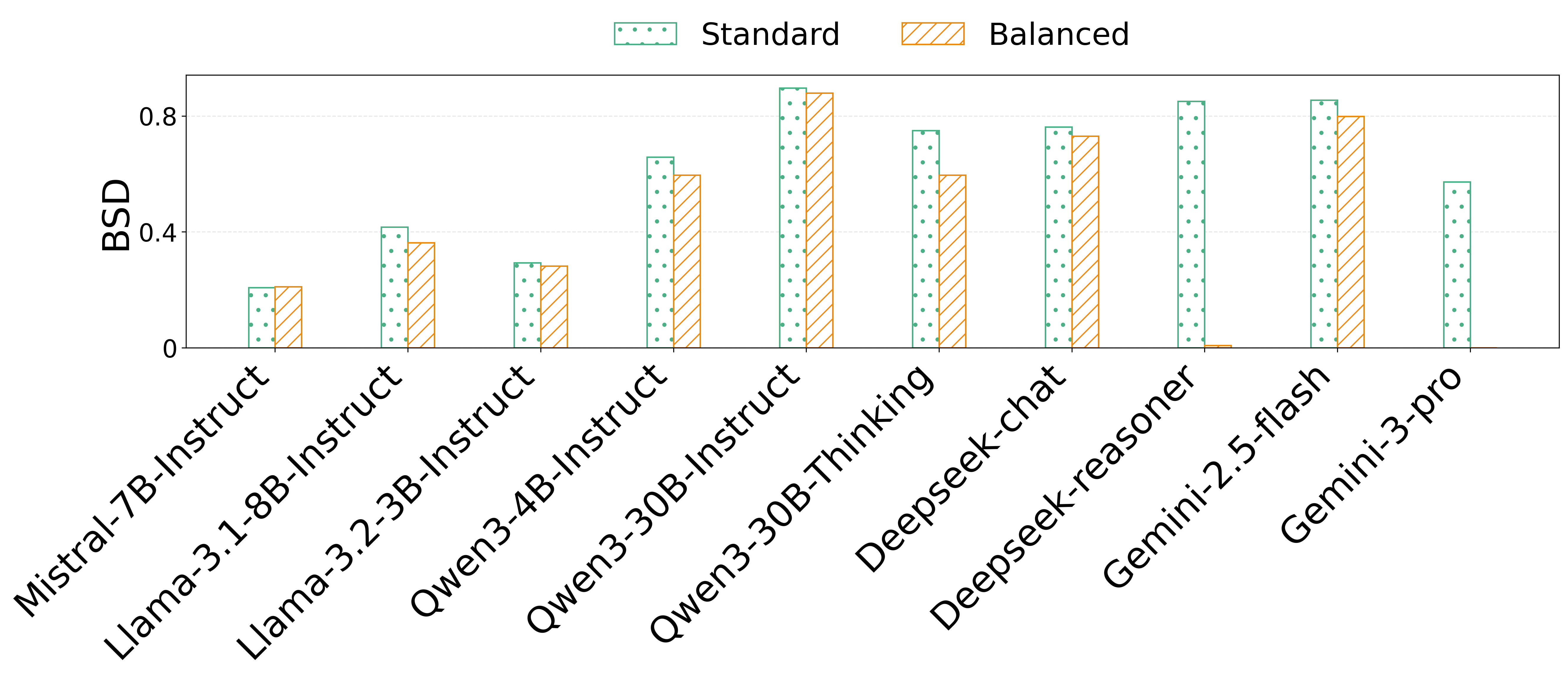}
}
\subfigure[Task 2: Distribution (standard vs. balanced)]{
    \includegraphics[width=0.45\textwidth]{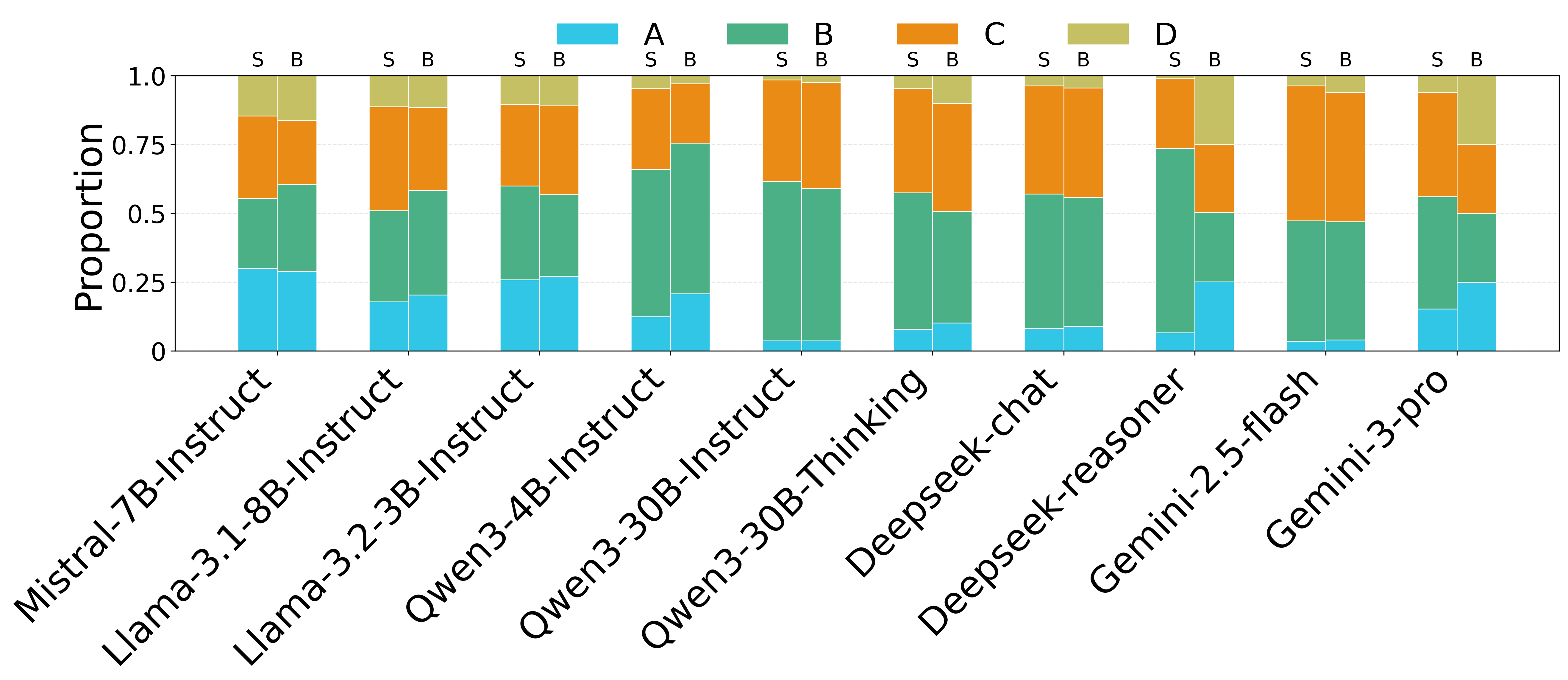}
}

\subfigure[Task 3: BSD (standard vs. balanced)]{
    \includegraphics[width=0.45\textwidth]{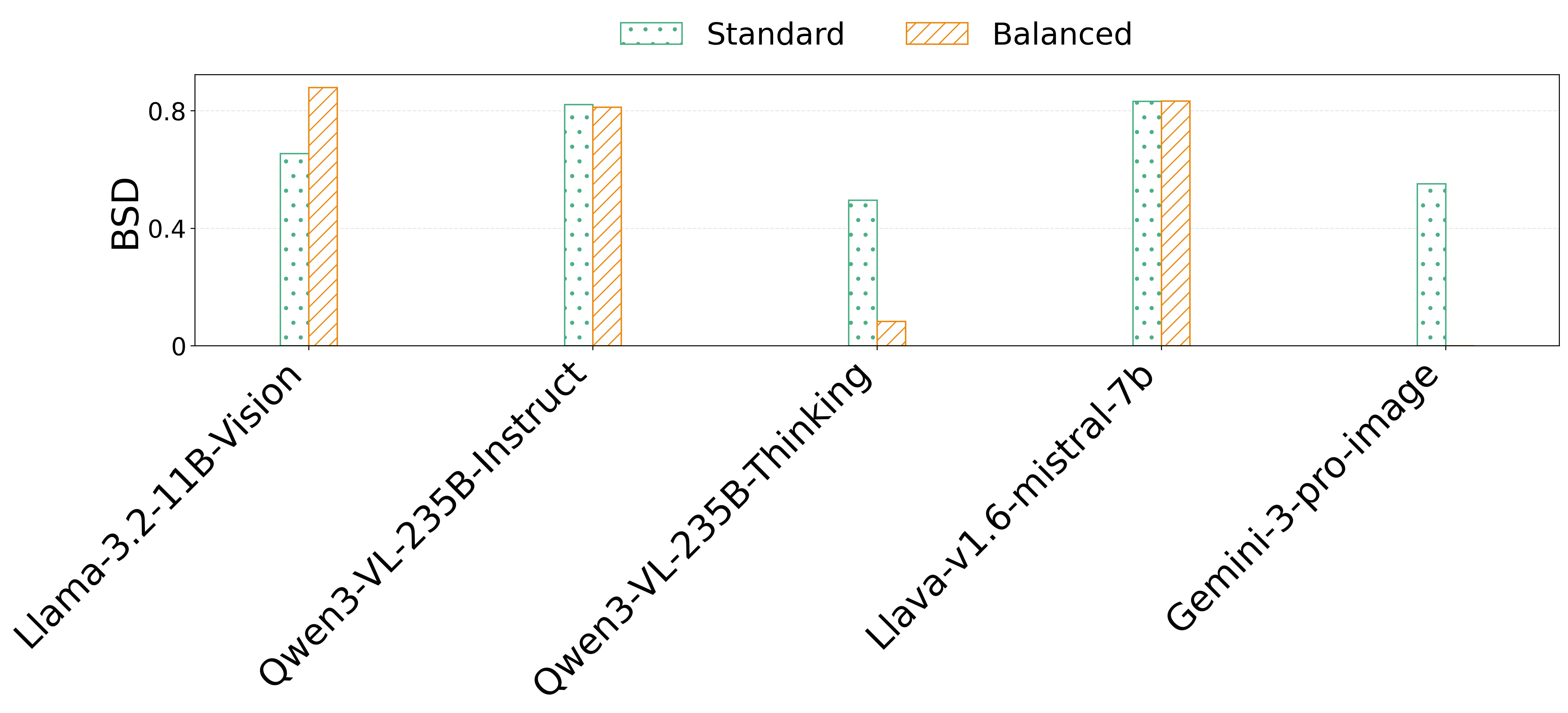}
}
\subfigure[Task 3: Distribution (standard vs. balanced)]{
    \includegraphics[width=0.45\textwidth]{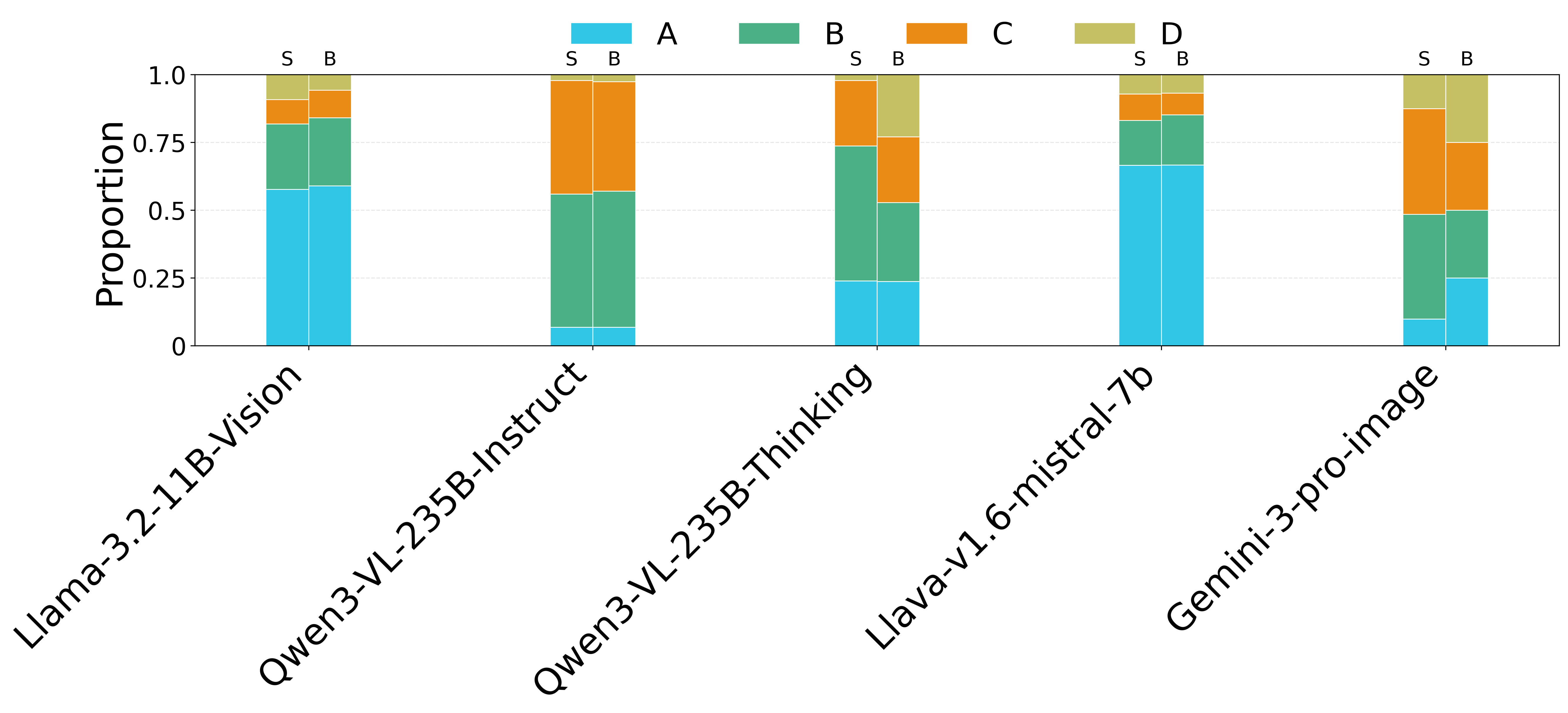}
}
\caption{Effect of balanced prompting on position bias across Task 2 and 3. }
\label{Task2-3_normal_balance}
\end{figure}

We present the results for Task 2 and Task 3 in Figure~\ref{Task2-3_normal_balance}. 

Consistent with Task 1, balanced prompting reduces position bias and leads to more uniform answer distributions across models.

\begin{figure}
\centering
\subfigure[Avg F1 (Standard)]{
    \includegraphics[width=0.3\textwidth]{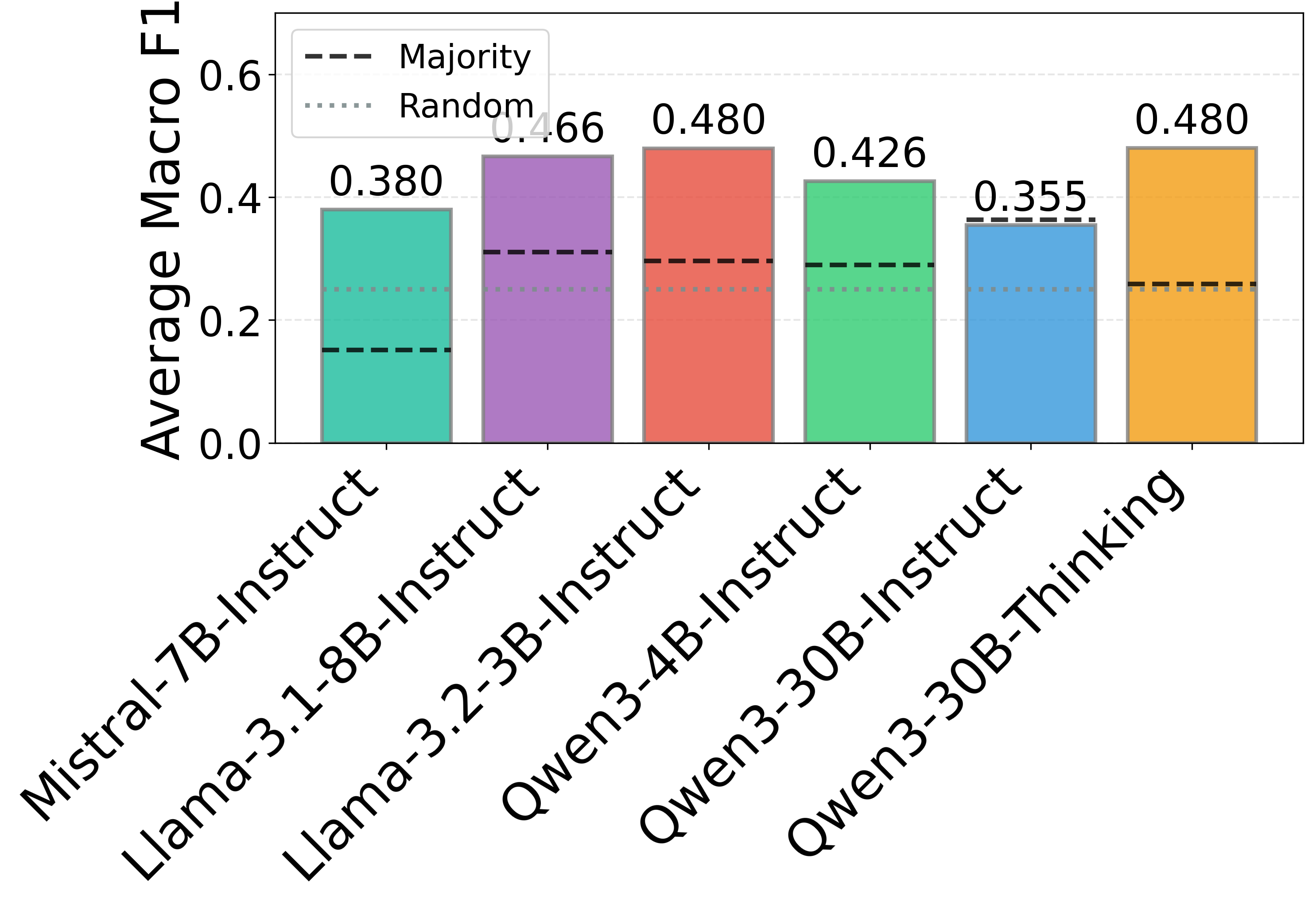}
}
\subfigure[Peak F1 (Standard)]{
    \includegraphics[width=0.3\textwidth]{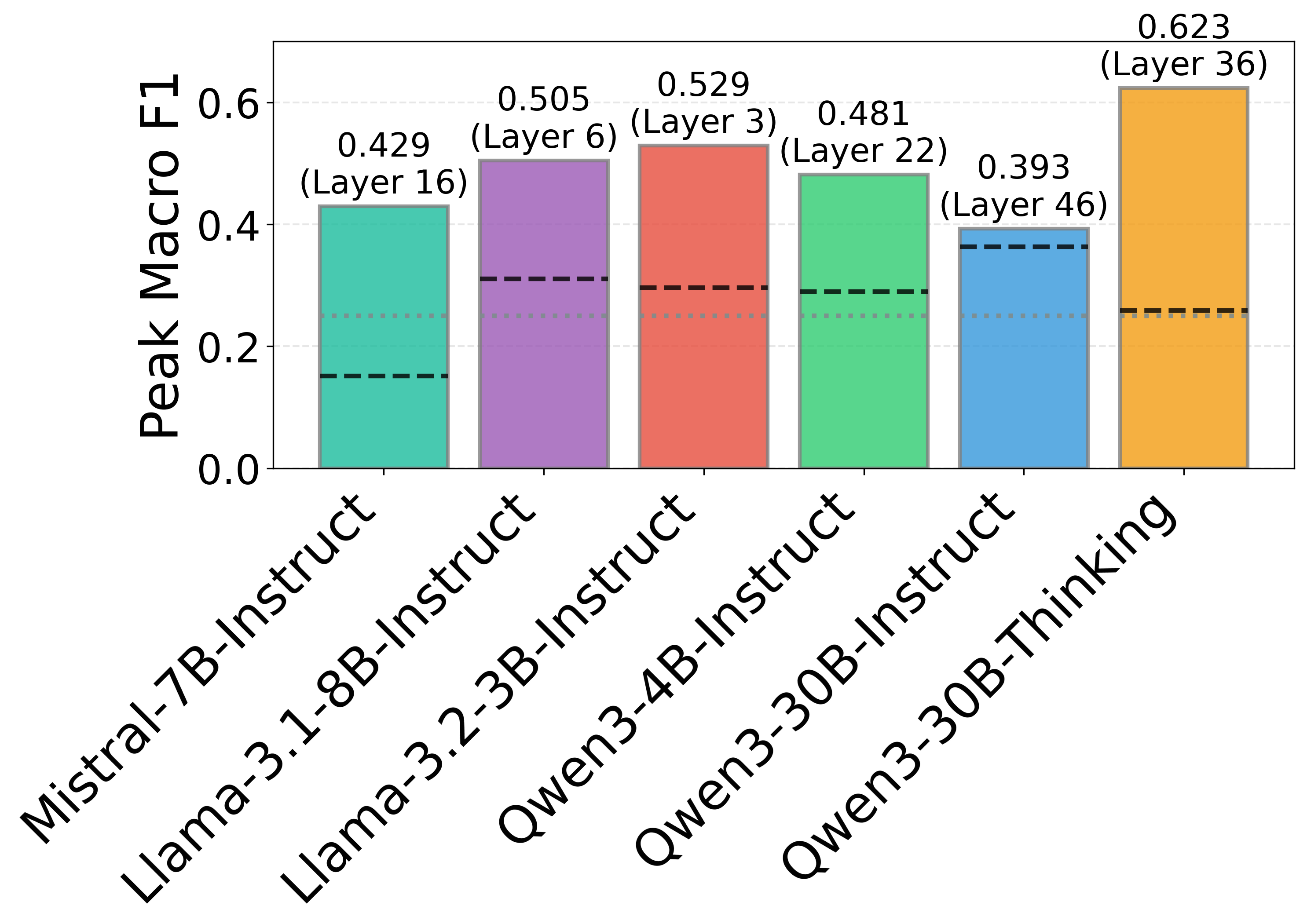}
}

\subfigure[Avg F1 (Standard w/o ID)]{
    \includegraphics[width=0.3\textwidth]{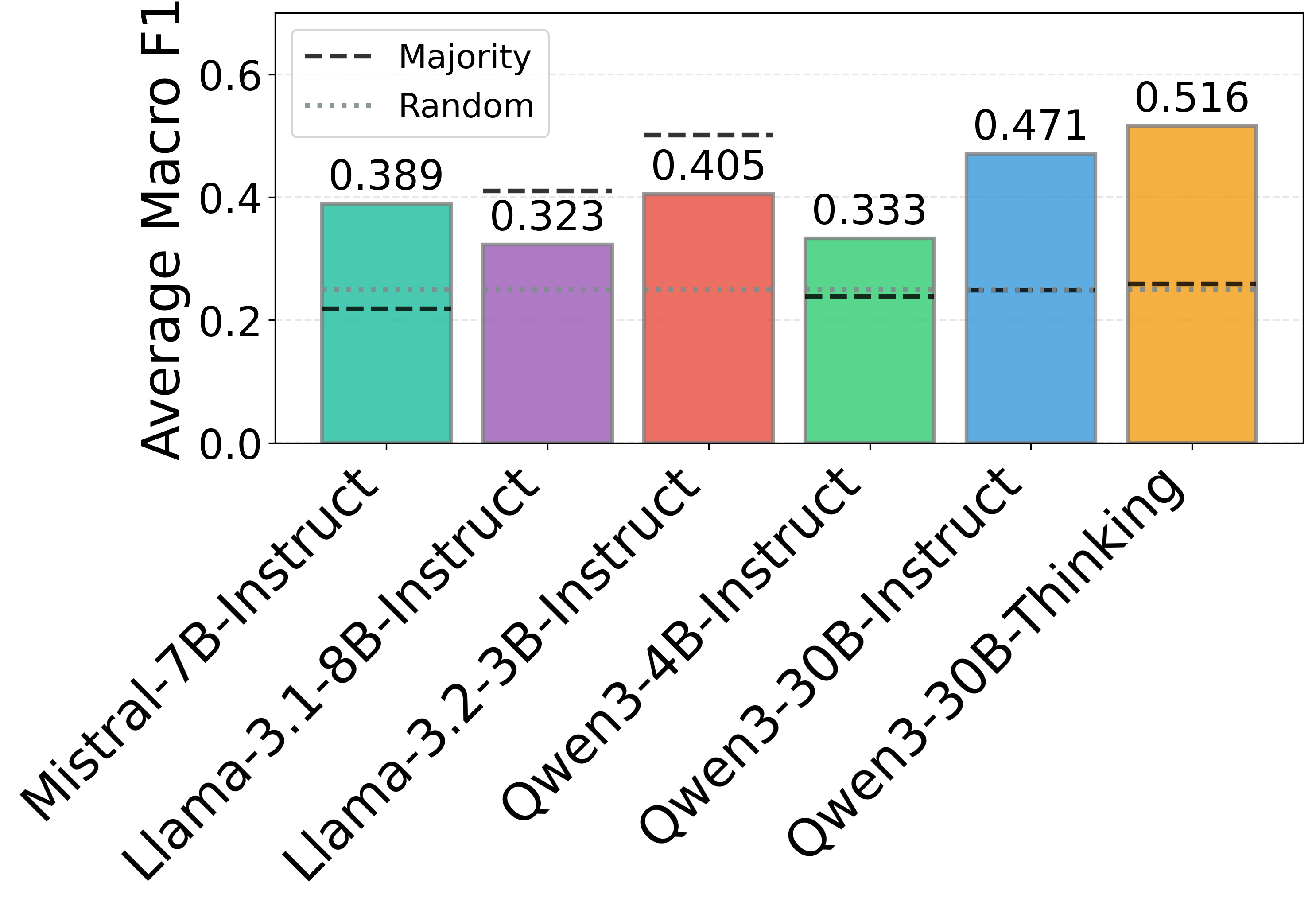}
}
\subfigure[Peak F1 (Standard w/o ID)]{
    \includegraphics[width=0.3\textwidth]{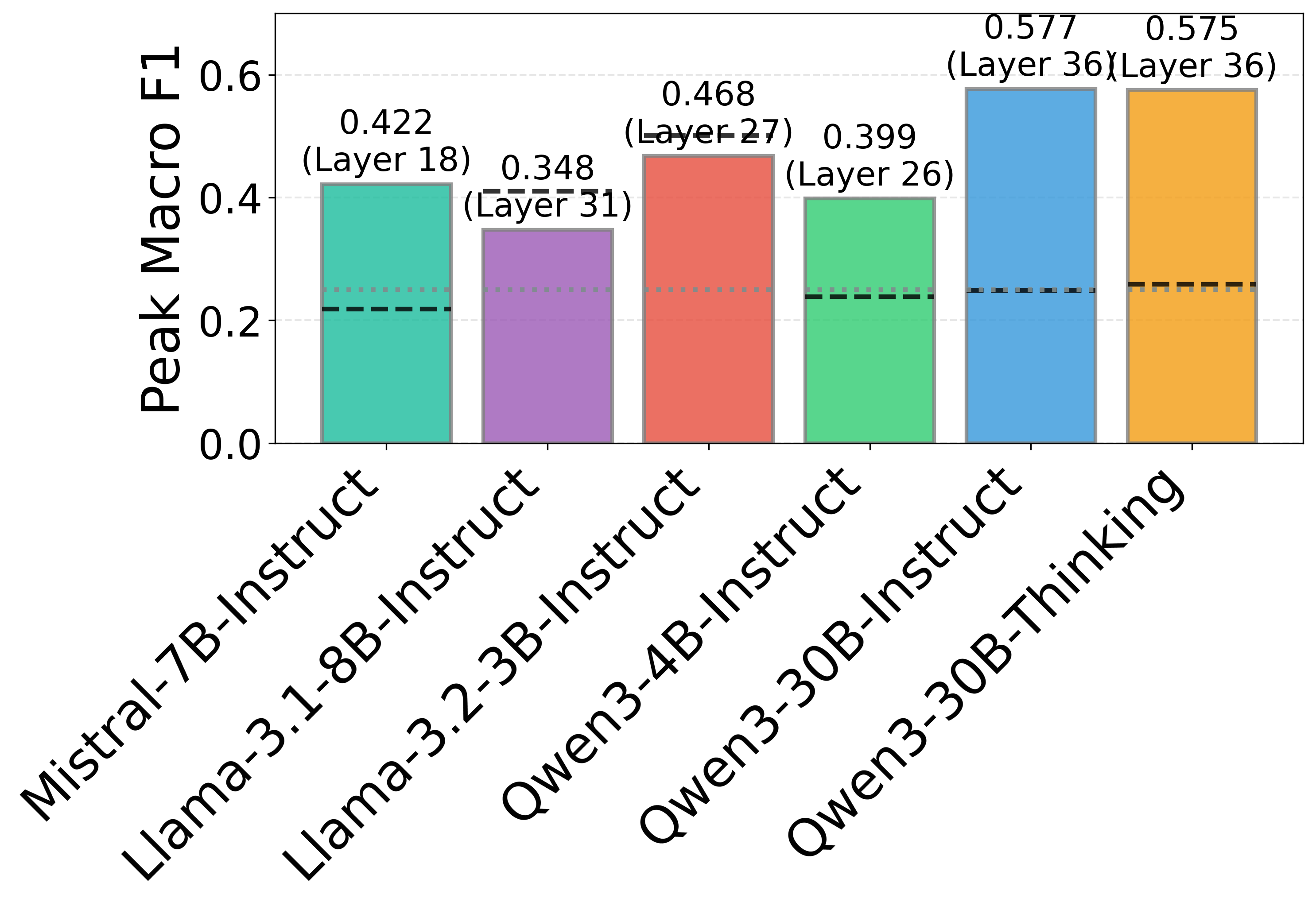}
}

\subfigure[Avg F1 (Balanced w/o ID)]{
    \includegraphics[width=0.3\textwidth]{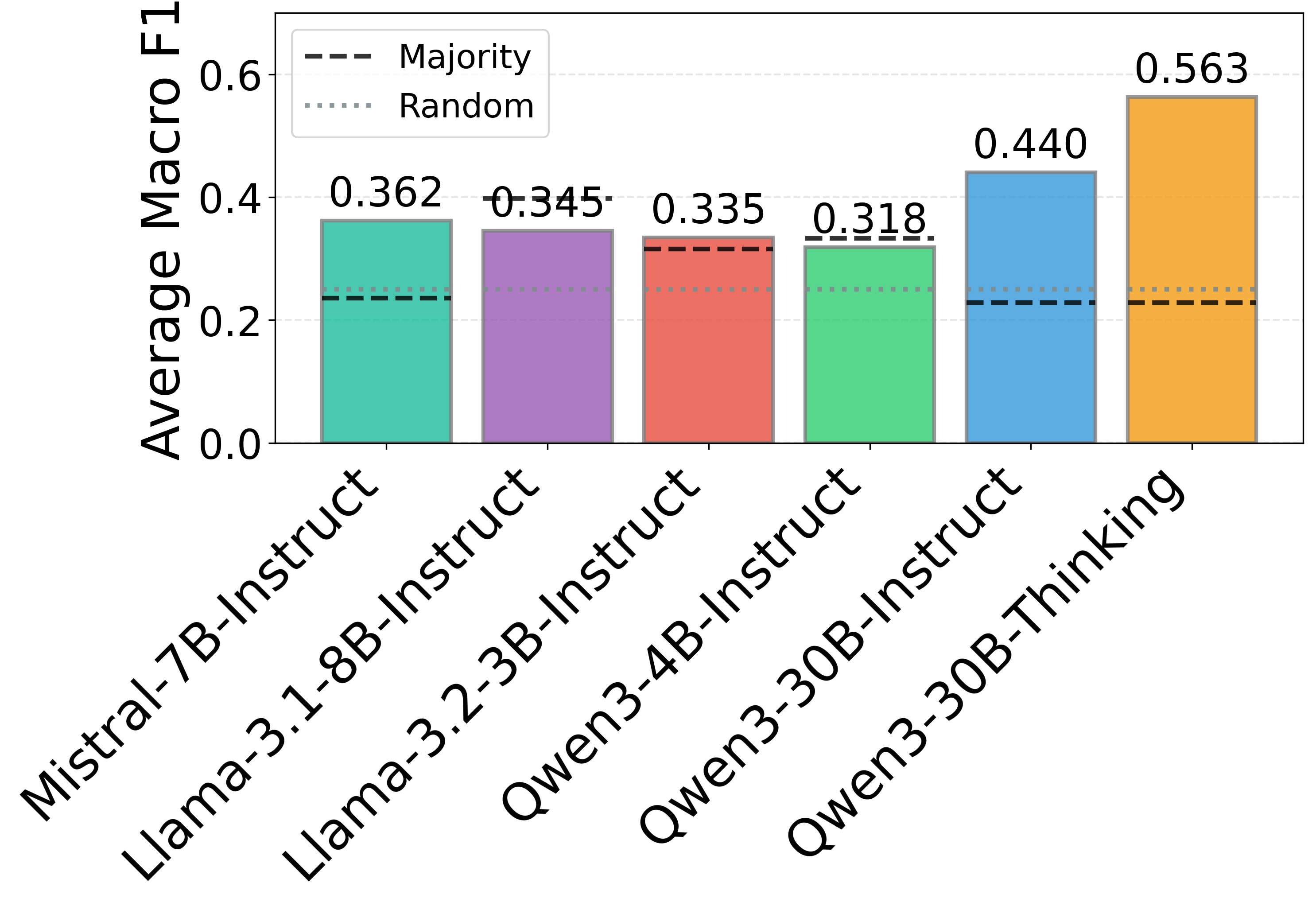}
}
\subfigure[Peak F1 (Balanced w/o ID)]{
    \includegraphics[width=0.3\textwidth]{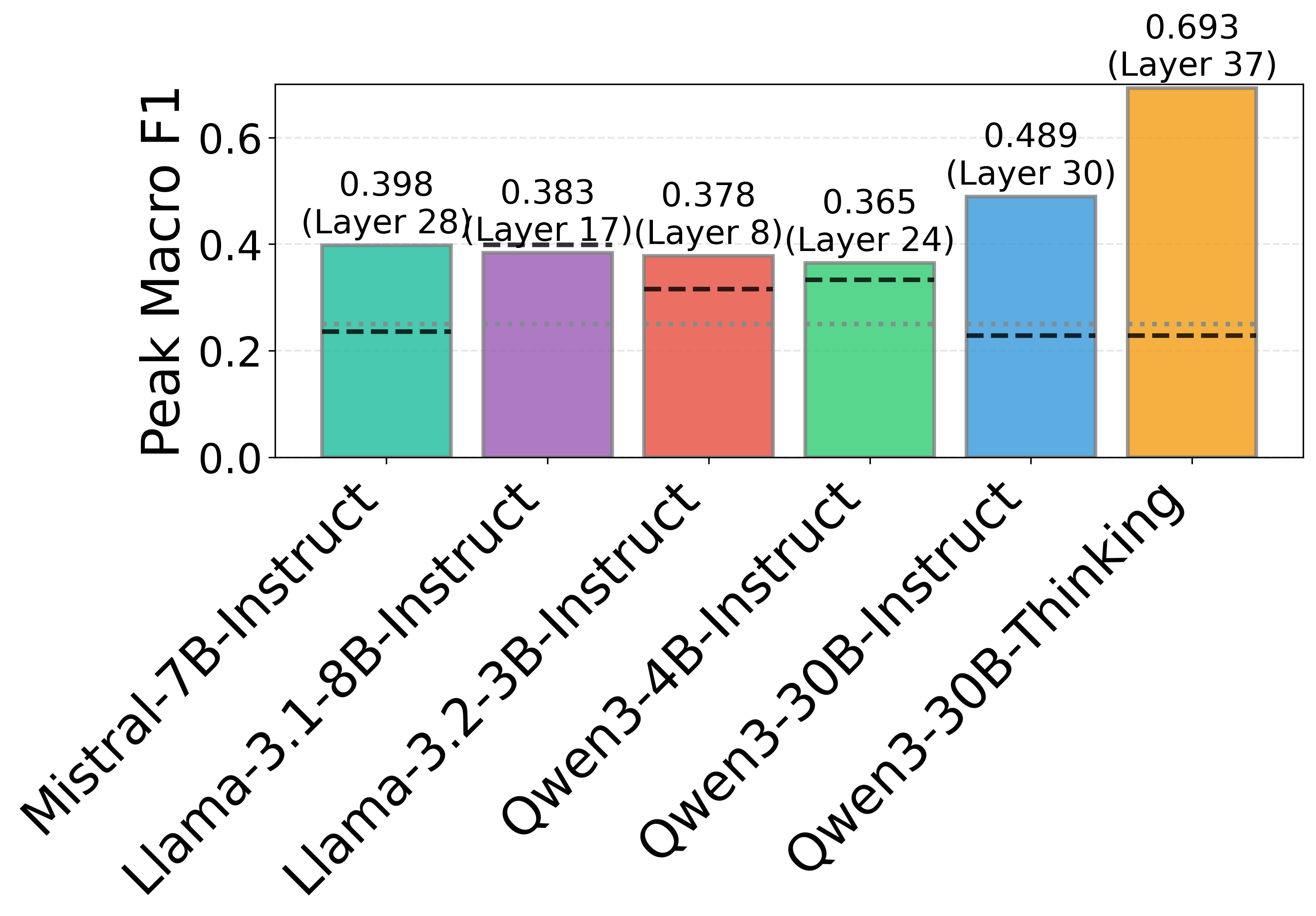}
}
\caption{Average and peak macro F1 scores of MLP probes for predicting correct answer positions. Results are shown under standard, standard (w/o ID) and balanced (w/o ID) prompting settings. }
\label{aver_peak_3}
\end{figure}

\section{Additional Results for Probing Experiments}
\label{appendix_probing}
Figure~\ref{aver_peak_3} presents the average and peak macro-F1 scores of MLP probes for predicting correct answer positions under different prompting settings. 

Under the standard prompting setting (a–b), most models achieve moderate probing performance, with peak F1 scores substantially higher than their averages, indicating that position information is localized in specific layers. Notably, reasoning-oriented or larger models (e.g., Qwen-30B-Thinking) achieve the highest peak performance, suggesting stronger and more explicit encoding of positional signals.

When removing explicit identifiers (standard w/o ID, c–d), performance drops for several models, particularly in average F1, indicating that explicit option markers contribute to more stable and globally distributed position representations. However, the presence of relatively high peak F1 scores implies that positional information is still internally encoded even without identifiers.

Under balanced prompting w/o ID (e–f), we observe a divergence across models. Stronger models (e.g., Qwen-30B variants) maintain or even improve both average and peak F1, while smaller models show limited gains or slight degradation. This suggests that while prompt-level balancing can reshape output distributions, its effect on internal representations depends on model capacity.

Overall, these results indicate that (1) answer position information is implicitly encoded in intermediate representations, (2) such encoding is partially influenced by prompt structure (e.g., identifiers), and (3) stronger models exhibit more robust and concentrated positional representations, especially in specific layers.

\section{Effect of Different Steering Positions and $\alpha$}
\label{appendix_alpha}

We report change rates across 27 layers of Llama3.2-3B-Instruct under different steering positions and $\alpha$.

\begin{figure}
    \vspace{-3mm}
    \centering
    \includegraphics[width=0.9\textwidth]{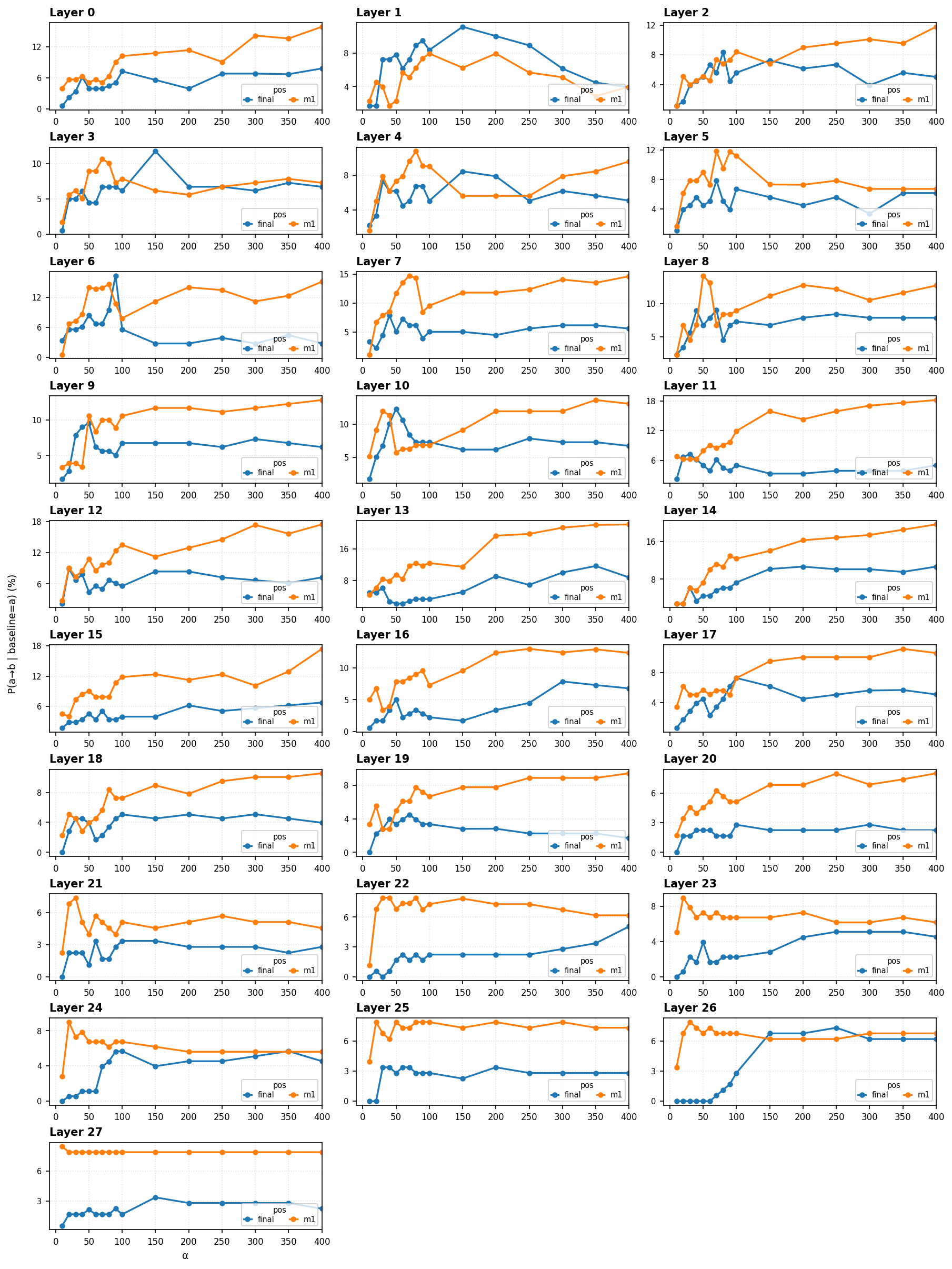}
\caption{A$\rightarrow$B change rate across 27 layers under mean-difference intervention over $\alpha$.}
    \label{mean_a_b}
\end{figure}

Figure~\ref{mean_a_b} shows the A$\rightarrow$B change rate across layers under mean-difference intervention with varying $\alpha$. Overall, the steering effect generally increases with $\alpha$, especially for the penultimate token, indicating that stronger intervention leads to more pronounced shifts. Across layers, intermediate layers (around layers 10–15) exhibit the strongest and most stable effects, while early and late layers show weaker or less consistent responses. In contrast, interventions at the final token remain relatively limited across all layers. These results suggest that intermediate representations are more sensitive to steering and that earlier intervention (at the penultimate token) allows the modification to more effectively influence the final output.

\begin{figure}
    \vspace{-3mm}
    \centering
    \includegraphics[width=0.9\textwidth]{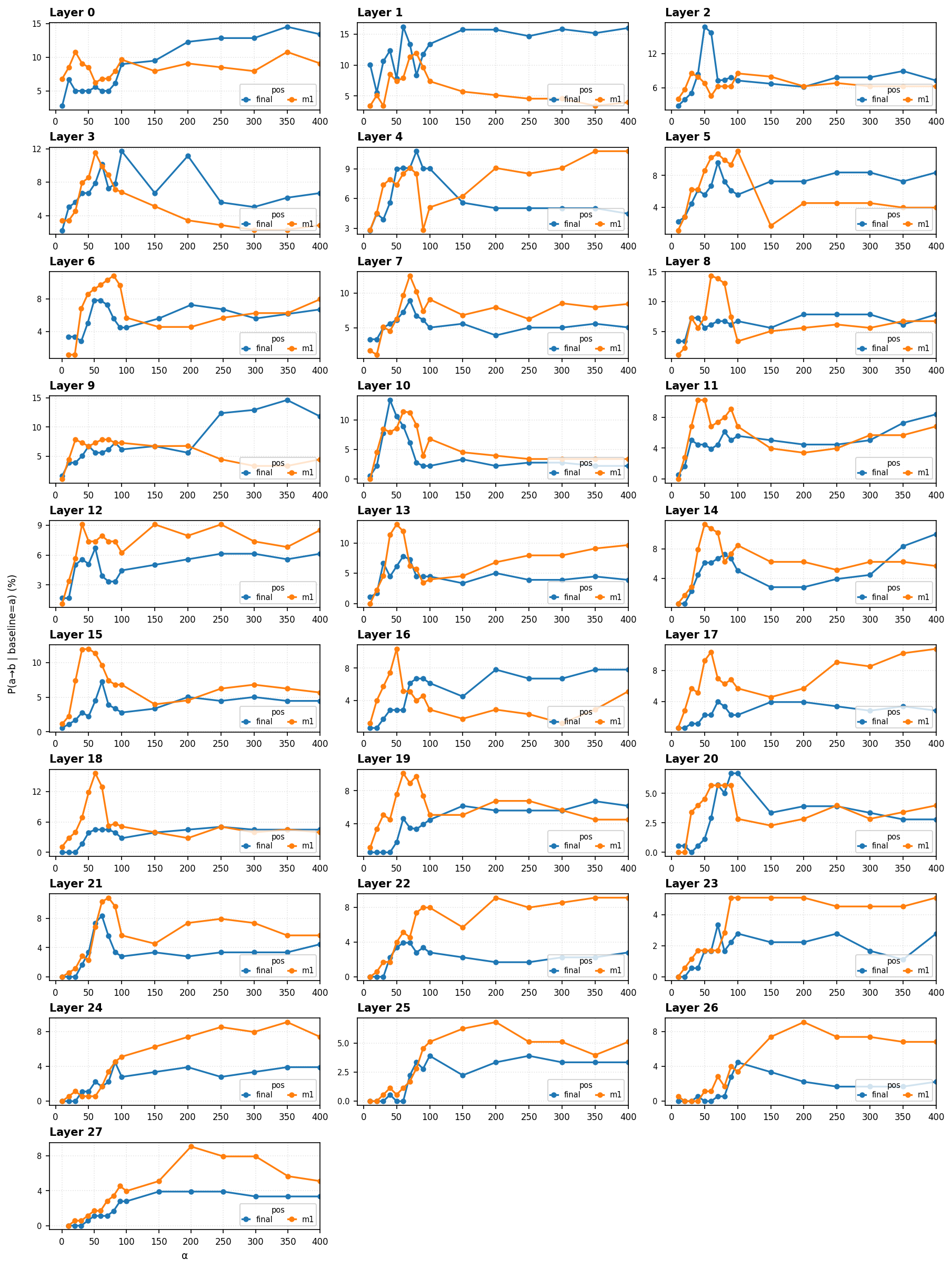}  
\caption{A$\rightarrow$B change rate across 27 layers under classifier-based intervention over $\alpha$.}
    \label{probe_a_b}
\end{figure}

Figure~\ref{probe_a_b} shows the A$\rightarrow$B change rate across layers under classifier-based intervention with varying $\alpha$. Compared to the mean-difference method, the steering effect here is generally weaker and less stable. While some early and middle layers exhibit noticeable peaks at small $\alpha$, the effect does not consistently increase with larger $\alpha$, and often plateaus or even decreases. Moreover, the difference between interventions at the penultimate and final tokens is less pronounced, indicating limited sensitivity to injection position. Overall, these results suggest that classifier-based directions provide less robust and less controllable steering signals, and their effectiveness varies more across layers and intervention strengths.

\section{Impact of Probe Capacity (MLP Hidden Size)}
\label{probe_size}

\begin{wrapfigure}{r}{0.3\textwidth}
\centering
\vspace{-5mm}
\includegraphics[width=0.3\textwidth]{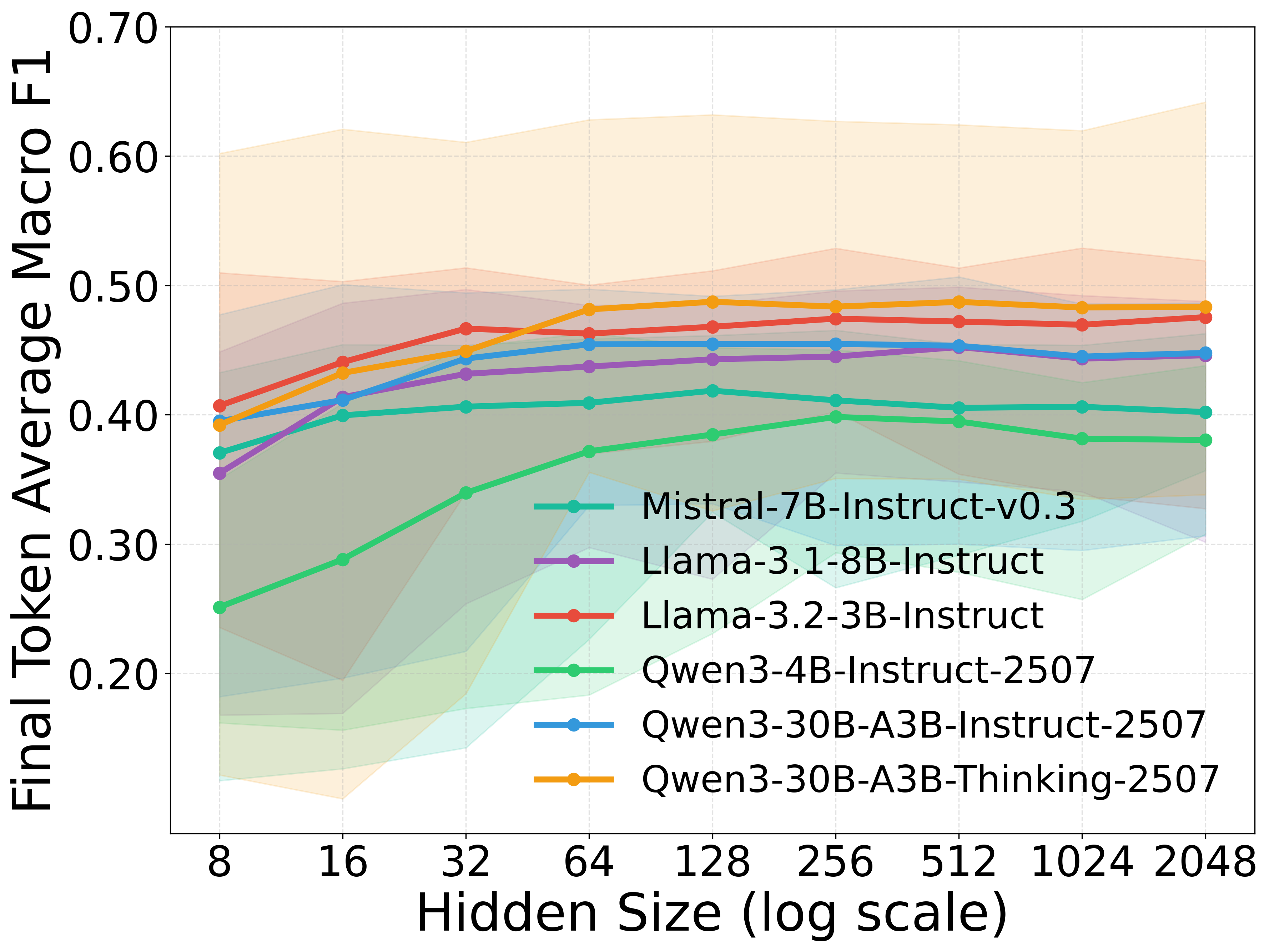}
\caption{Macro-F1 scores under different MLP hidden sizes on Task 1, using balanced prompting and last-token representations.}
\label{last_token_mlp}
\vspace{-3mm}
\end{wrapfigure}

We further examine the impact of probe capacity by varying the hidden size of the MLP and measuring performance. As shown in Figure~\ref{last_token_mlp}, performance plateaus before hidden size 256 across models, while larger sizes may lead to overfitting. This suggests that position-related signals are relatively salient and can be captured with low-capacity probes.

\section{Target vs. Off-target Trade-off at the Final Token}
\label{appendix_final_token_tradeoff}

\begin{wrapfigure}{r}{0.3\textwidth}
\centering
\includegraphics[width=0.3\textwidth]{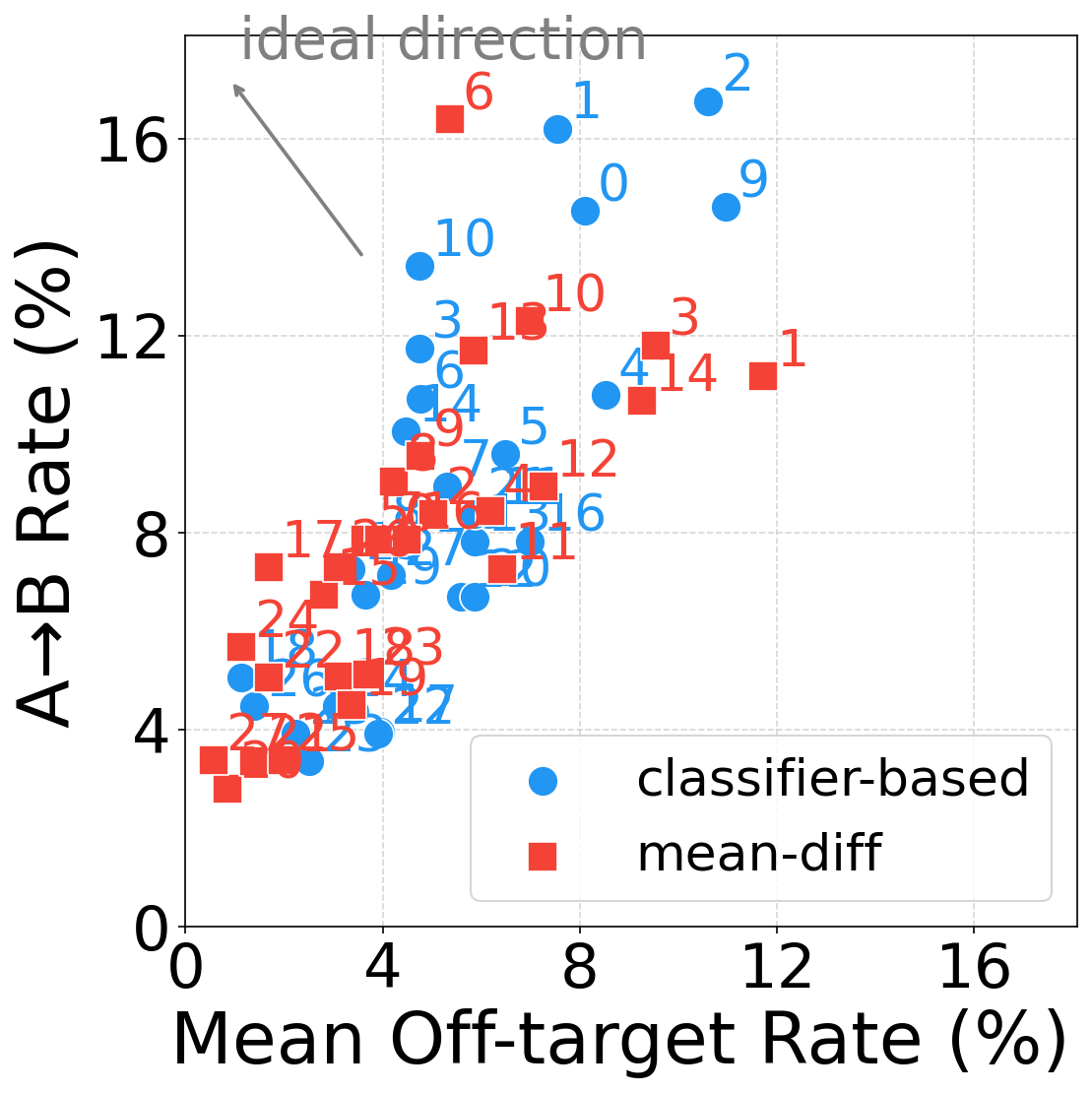}
\caption{Target vs. off-target trade-off at the final token. Each point corresponds to a layer with its optimal intervention strength.}
\label{final_target_off}
\vspace{-3mm}
\end{wrapfigure}

At the final token, layer 6 achieves relatively high target shifts (~16\%) with moderate off-target changes (~6\%). In contrast, classifier-based steering attains similar target shifts in early layers but incurs substantially higher off-target changes (>10\%), indicating less precise control.

\section{Additional Case Study}
\label{appendix_bad_case}
Table~\ref{bad_case} presents several representative failure cases observed under steering interventions.
We identify three common error patterns.

First, \textbf{label flips} occur when the intervention changes the predicted option identifier while leaving the semantic content of the question and options unchanged. In the first example, the model originally selects the correct answer ``a) heat of fusion'', but after steering it shifts to ``b) latent heat'', suggesting that the intervention perturbs the mapping between internal confidence and option selection rather than improving semantic reasoning.

Second, \textbf{semantic incoherence} arises when steering disrupts the consistency between the question and the generated options. In the second example, the question asks about the world’s largest island, but the intervened output replaces island candidates with ocean names, indicating that strong interventions can corrupt task-specific semantic constraints and lead to topic drift.

Third, \textbf{incorrect reasoning} reflects cases where the model preserves the question format but fails to maintain logically valid reasoning. In the trigonometric example, the intervened model generates options inconsistent with the equation $\cos(q)=0.5$ and selects an incorrect answer, suggesting that steering may distort latent reasoning trajectories even when surface-level structure remains intact.

In addition, we observe a fourth type of failure where steering locally corrupts token generation, producing abnormal or nonsensical symbols at the intervention position while subsequent tokens remain largely coherent. This phenomenon suggests that strong activation perturbations can destabilize the representation of the current decoding step without fully disrupting the model’s longer-range contextual dynamics. In other words, the intervention may temporarily push the hidden state outside the model’s typical representation manifold, leading to malformed tokens or unexpected characters, after which the autoregressive generation process partially recovers at later positions. This behavior further indicates that activation steering can introduce low-level decoding instability in addition to higher-level semantic and reasoning errors.

Overall, these failure cases suggest that activation steering does not merely bias answer positions, but can also interfere with semantic alignment, reasoning consistency, and decoding stability. These observations highlight the importance of carefully controlling intervention strength and injection location to avoid degrading the model’s underlying task representations.

\begin{table}[t]
\centering
\small
\caption{Representative failure cases under steering.}
\label{bad_case}
\begin{tabular}{p{0.28\linewidth} p{0.28\linewidth} p{0.16\linewidth} p{0.1\linewidth}}
\toprule
\textbf{Baseline} & \textbf{Intervened} & \textbf{Type} & \textbf{Correctness} \\
\midrule

\parbox[t]{\linewidth}{
\raggedright
\textbf{Q:} What is the term for the amount of heat energy released when a substance melts? \\
\textbf{Options:} a) heat of fusion b) latent heat c) enthalpy change d) specific heat capacity \\
\textbf{Answer:} a
}
&
\parbox[t]{\linewidth}{
\raggedright
\textbf{Q:} What is the term for the amount of heat energy released when a substance melts? \\
\textbf{Options:} a) heat of fusion b) latent heat c) enthalpy change d) specific heat capacity \\
\textbf{Answer:} b
}
&
Label Flip & $\times$ \\

\midrule

\parbox[t]{\linewidth}{
\raggedright
\textbf{Q:} What is the world's largest island? \\
\textbf{Options:} a) Greenland b) Borneo c) New Guinea d) Madagascar \\
\textbf{Answer:} a
}
&
\parbox[t]{\linewidth}{
\raggedright
\textbf{Q:} What is the world's largest island? \\
\textbf{Options:} a) Arctic Ocean b) Southern Ocean c) Indian Ocean d) Pacific Ocean \\
\textbf{Answer:} b
}
&
Semantic Incoherence & $\times$ \\

\midrule

\parbox[t]{\linewidth}{
\raggedright
\textbf{Q:} Solve for $q$: $\cos(q) = 0.5$ \\
\textbf{Options:} a) $30^\circ$ b) $45^\circ$ c) $60^\circ$ d) $90^\circ$ \\
\textbf{Answer:} a
}
&
\parbox[t]{\linewidth}{
\raggedright
\textbf{Q:} Solve for $q$: $\cos(q) = 0.5$ \\
\textbf{Options:} a) $90^\circ$ b) $180^\circ$ c) $270^\circ$ d) $360^\circ$ \\
\textbf{Answer:} b
}
&
Incorrect Reasoning & $\times$ \\

\bottomrule
\end{tabular}
\end{table}

\section{Task Prompts}

This section provides the prompts used for the three experimental tasks. All prompts are applied consistently across models and settings unless otherwise specified.

\subsection{Task 1}

\begin{verbatim}
Please generate 40 multiple-choice questions related to the discipline {disc} and the
topic {topic}. Only generate questions within the specified discipline and topic.

Each question must satisfy the following requirements:

1. Each question includes exactly one correct answer and three plausible distractors.
2. Each question must be formatted in the following JSON structure:

[
  {
    "question": "Question text",
    "options": {
      "a": "Option A",
      "b": "Option B",
      "c": "Option C",
      "d": "Option D"
    },
    "answer": "correct option text"
  }
]

3. The "answer" field must exactly match one of the four option values.
4. Questions should cover a mix of easy, medium, and hard difficulty levels.
5. Do not include any explanations or additional text outside the JSON array.

Strict constraints:
- Exactly 40 questions must be generated.
- Each question must include exactly 4 options (a–d).
- The output must be valid JSON.
\end{verbatim}

\subsection{Task 2}

\begin{verbatim}
[Notice]
All input articles are from publicly available BBC News content and are used solely for
academic research purposes, including natural language processing evaluation and reading 
comprehension dataset generation. The content will not be used for any commercial purpose
or redistributed.

[Task]
Based on the following article, generate {q_count} different reading comprehension 
questions, each with 4 options (a, b, c, d), and number each question (1, 2, 3, …).

Requirements:
- The distractor options should be incorrect but plausible distractors.
- All questions must be answerable from the following article.
- Order the questions from easiest to hardest.
- Questions should cover the main information of the article, including facts, details,
people, and events.
- Output json format example:

[{
  "question": "Question text",
  "options": {
    "a": "Option A",
    "b": "Option B",
    "c": "Option C",
    "d": "Option D"
  },
  "answer": ""
}]

\end{verbatim}

\subsection{Task 3}

\begin{verbatim}
You are given **one image** and **a short textual description** of the same image.

Your task is to generate **{count} multiple-choice visual question answering (VQA) 
items** based on the image and the description.

#### **Question requirements**

* Each question must be answerable using the image content (the description can be used 
as support).
* You must rely on visual evidence from the image. Do not generate questions that can 
be answered from the description alone.
* Questions should cover **diverse aspects** of the image, including:

  * objects and entities
  * attributes (color, number, shape, material)
  * spatial relationships
  * actions or interactions
  * scene or environment
  * fine-grained or local details
* Avoid yes/no questions.
* Do not include external knowledge.

#### **Answer requirements**

* Each question must have **exactly four options**: A, B, C, D.
* **Only one option is correct** for each question.
* Do **not** reveal the correct option inside the question text.

#### **Output format**

Output a JSON array with the following structure:

[
  {
    "question": "Question text here",
    "options": {
      "A": "Option A",
      "B": "Option B",
      "C": "Option C",
      "D": "Option D"
    },
    "answer": "correct option"
  }
]

**Important Constraints (MUST follow):**
* The response MUST include exactly {count} questions.
* Each question MUST include exactly 4 options.
* Each question MUST include an "answer" field.
* The answer MUST be exactly one of the four options.
* Do NOT omit any field. Missing fields will be considered invalid output.

**Output Check:**
Before outputting, ensure that:
* Every question has 4 options.
* Every question has 1 answer.
* The JSON format is complete and valid.

\end{verbatim}



\end{document}